\documentclass[conference]{IEEEtran}
\usepackage{times}

\usepackage[numbers]{natbib}
\usepackage{amsmath}
\usepackage{amssymb}
\usepackage{amsfonts}
\usepackage{graphicx}
\usepackage{xcolor}
\usepackage{booktabs}
\usepackage{multirow}
\usepackage{multicol}
\usepackage{makecell}
\usepackage{subcaption}
\let\labelindent\relax
\usepackage{enumitem}
\usepackage{colortbl}
\definecolor{colorblue}{RGB}{23,105,255}
\usepackage[colorlinks=true, citecolor=colorblue, linkcolor=colorblue, urlcolor=colorblue]{hyperref}

\begin{document}

\title{One Hand to Rule Them All: Canonical Representations for Unified Dexterous Manipulation}


\author{
  \textbf{Zhenyu Wei} \quad \textbf{Yunchao Yao} \quad \textbf{Mingyu Ding} \\
  University of North Carolina at Chapel Hill
}

\twocolumn[{%
\renewcommand\twocolumn[1][]{#1}%
\maketitle
\begin{center}
  \vspace{-16pt}
  \includegraphics[width=\linewidth]{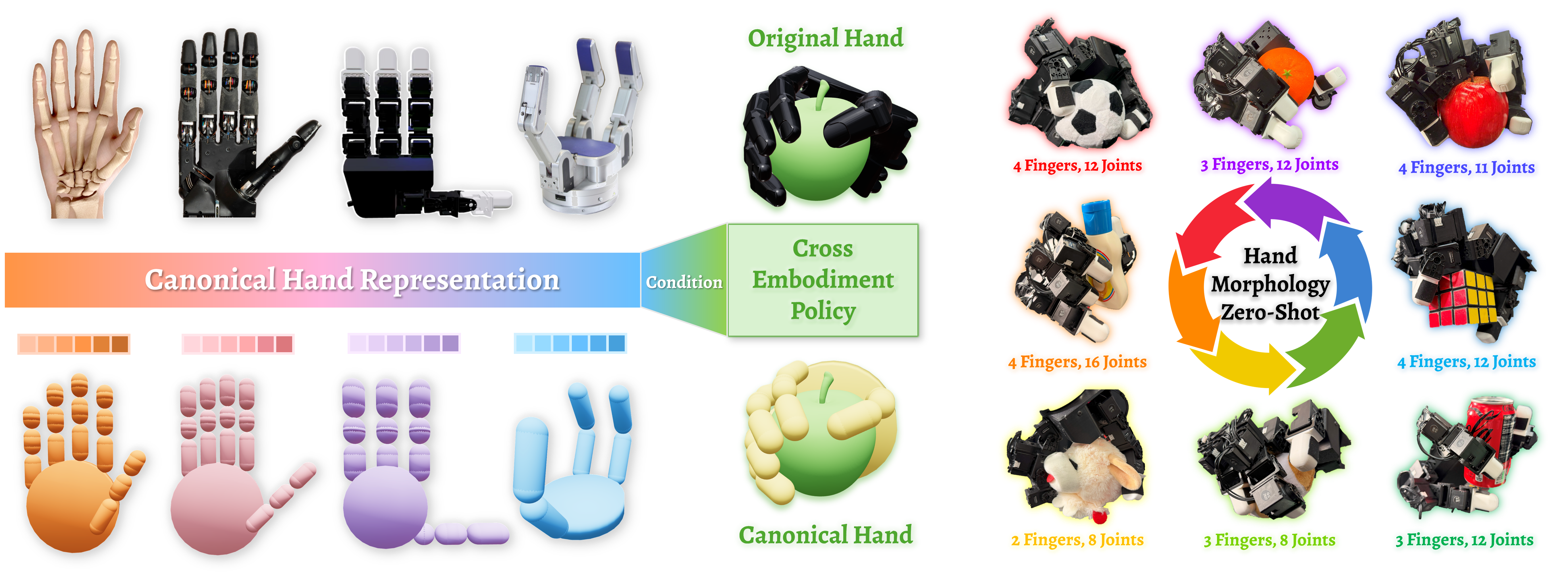}
  \vspace{-16pt}
  \captionof{figure}{We introduce a canonical hand representation that unifies diverse dexterous hands into a shared parameter space and canonical URDF format, serving as a condition for cross-embodiment policy learning. It enables dexterous grasping and zero-shot generalization to novel hand morphologies, highlighting its potential for a wide range of dexterous manipulation tasks.}
  \label{fig:teaser}
\end{center}%
}]

\begin{abstract}
Dexterous manipulation policies today largely assume fixed hand designs, severely restricting their generalization to new embodiments with varied kinematic and structural layouts. 
To overcome this limitation, we introduce a parameterized canonical representation that unifies a broad spectrum of dexterous hand architectures. It comprises a unified parameter space and a canonical URDF format, offering three key advantages.
1) The parameter space captures essential morphological and kinematic variations for effective conditioning in learning algorithms.
2) A structured latent manifold can be learned over our space, where interpolations between embodiments yield smooth and physically meaningful morphology transitions.
3) The canonical URDF standardizes the action space while preserving dynamic and functional properties of the original URDFs, enabling efficient and reliable cross-embodiment policy learning.

We validate these advantages through extensive analysis and experiments, including grasp policy replay, VAE latent encoding, and cross-embodiment zero-shot transfer.
Specifically, we train a VAE on the unified representation to obtain a compact, semantically rich latent embedding, and develop a grasping policy conditioned on the canonical representation that generalizes across dexterous hands.
We demonstrate, through simulation and real-world tasks on unseen morphologies (\emph{e.g.}, 81.9\% zero-shot success rate on 3-finger LEAP Hand), that
our framework unifies both the representational and action spaces of structurally diverse hands, providing a scalable foundation for cross-hand learning toward universal dexterous manipulation.
Project Page: \url{https://zhenyuwei2003.github.io/OHRA/}

\end{abstract}

\IEEEpeerreviewmaketitle

\section{Introduction}
\label{sec:intro}

\begin{figure*}[t]
  \centering
  \includegraphics[width=\textwidth]{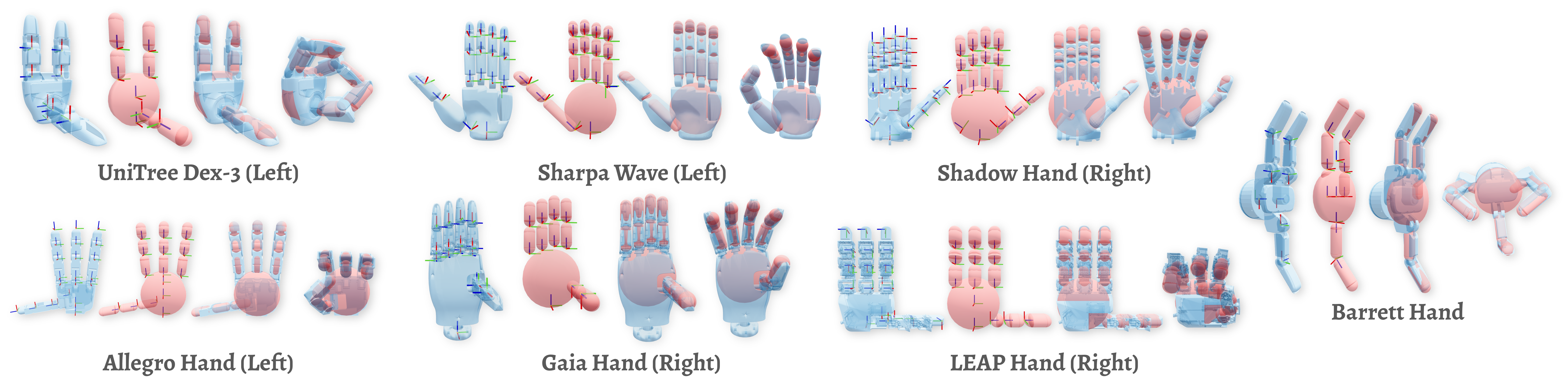}
  \vspace{-10pt}
  \caption{Comparison of canonical and original URDFs across seven dexterous hands with different finger numbers and handedness. For each hand (from left to right): canonical URDF, original URDF, overlay of initial poses, and overlay of grasp poses, showing close morphological and kinematic consistency between the canonical and original models.}
  \vspace{-10pt}
  \label{fig:hand_vis}
\end{figure*}

Dexterous robotic hands with high degrees of freedom (DoF) and anthropomorphic designs offer advanced flexibility and control beyond the capabilities of simpler parallel or underactuated grippers~\cite{34763,897777}, enabling robots to interact with diverse objects in dynamic and unstructured environments.
Recent advances in both learning-based and model-based approaches have driven impressive progress in object grasping~\cite{wei2024dro,li2022gendexgrasp,wang2022dexgraspnet,zhang2024dexgraspnet,wan2023unidexgrasp++,shao2020unigrasp}, dynamic in-hand manipulation~\cite{chen2023visual, chen2022system, wang2024lessons, andrychowicz2020learning, yin2023rotating}, and tool use for goal-directed behaviors~\cite{agarwal2023dexterous, liang2025dexhanddiff, lei2025rl, ze20243d, rajeswaran2017learning, xu2025dexumi, guo2024telepreview}.
However, the specific DoF requirements and structural layouts vary substantially across tasks and robotic platforms.
As a result, existing methods generally remain tailored to specific robotic hands, limiting their generalization and reusability across embodiments. 

Variations in morphology, DoF, and kinematic layouts make policies trained on one hand design difficult to transfer to another~\cite{xu2025dexumi}. This lack of interoperability increases the cost of data collection and prevents leveraging heterogeneous datasets across platforms.
Recent studies have begun exploring cross-embodiment manipulation: $\mathcal{D(R,O)}$ Grasp~\cite{wei2024dro} transfers grasping skills across hands by modeling hand–object interactions, but remains task-specific;
DexUMI~\cite{xu2025dexumi} uses the human hand as a universal interface, yet its reliance on human-like kinematics and teleoperation setups limits scalability; and particle-based dynamics learning methods~\cite{he2025scaling} model both hands and objects as particle systems, enabling transfer for deformable manipulation but struggling with structurally distinct hands and broader manipulation skills.
Despite these efforts, a central challenge remains: \emph{How to establish a unified representation and action space that policies can generalize across different hand embodiments and tasks?}

Addressing this question requires solving two fundamental challenges: (i) obtaining a compact yet expressive representation of hand morphology that can serve as a conditioning input for learning-based models, 
and (ii) establishing a unified action space that allows policies to function seamlessly across hands with varying DoFs and kinematic structures.
Although the URDF format encodes a complete specification of hand geometry and kinematics, its hierarchical and heterogeneous nature makes it ill-suited for direct use in neural networks for learning-based approaches. This motivates us to introduce a parameterized canonical representation, which encodes hand structure in a learning-friendly format while standardizing the action space across diverse dexterous hands.

Building on this canonical representation, we encode the geometric and kinematic properties of dexterous hands into a fixed set of parameters defined under a canonical URDF format. This format can approximate a wide range of robotic hand designs while maintaining a unified action space, where inactive joints are treated as dummy variables.
Through an automated pipeline, existing hand URDFs are converted into this canonical format and corresponding parameter representation, yielding two unified spaces: (i) a parameter space that compactly captures hand morphology, and (ii) an action space that standardizes joint movements across embodiments. 
This unified formulation enables joint policy training across diverse hand embodiments, bridging their structural and dimensional discrepancies.

We validate our framework through three key experiments. 
First, training a variational autoencoder (VAE) on the canonical parameter space reveals a structured latent manifold for hand geometry and kinematics, where interpolation between different embodiments produces smooth, physically meaningful morphology transitions.
Second, grasp policy replay and in-hand reorientation tasks show that the canonical URDF preserves the kinematic and functional characteristics of the original hands, achieving performance comparable to their native URDFs. 
Finally, conditioning on hand morphology representations, we train a cross-embodiment dexterous grasping policy that generalizes across diverse hand designs. We first demonstrate effective policy transfer across three distinct robotic hands, where the shared policy outperforms per-hand baselines. We further scale this approach by training on near one hundred LEAP Hand variants, enabling zero-shot generalization to unseen hand morphologies and robust grasping performance in both simulation and real-world experiments, without additional fine-tuning.

Our main contributions are summarized as follows:
\begin{enumerate}[leftmargin=1.2em]
    \item We propose a canonical representation for dexterous hands that standardizes diverse morphologies and kinematic structures into a unified parameterized format, enabling consistent and learning-friendly structural encoding across hands.
    \item Extensive experiments including morphology latent interpolation, grasp policy replay, in-hand reorientation, and unseen cross-hand grasping demonstrate that the canonical format not only faithfully preserves functional behavior but also provides a unified action space for zero-shot and effective policy transfer across embodiments.
    \item For the first time, we establish an interpretable, interpolable, and scalable representation foundation that enables joint policy training for cross-embodiment dexterous manipulation, paving the way for unified large-scale and morphology-aware learning.
\end{enumerate}

\section{Related Works}
\label{sec:related}

\noindent \textbf{Dexterous Manipulation.}
High-DoF robotic hands provide the articulation needed for fine-grained contact control and multi-stage manipulation, supporting tasks from stable grasping to dynamic in-hand reconfiguration and tool-mediated interactions~\cite{huang2023dynamic, zhang2025catch, chen2022system, wei2024dro, li2022gendexgrasp, wang2024cyberdemo, xu2025dexsingrasp, liang2025dexhanddiff}. These capabilities have been extensively explored through analytic models and data-driven learning, producing strong performance on individual embodiments. In particular, reinforcement learning and imitation learning have yielded high-quality grasping and manipulation policies when confined to a fixed dexterous hand~\cite{yuan2024robot, chen2023visual, yuan2024learning, ze20243d}. However, because such policies are optimized for the specific kinematics, actuation, and workspace of a single hand, they become tightly coupled to that embodiment~\cite{xu2023unidexgrasp, wan2023unidexgrasp++}. This embodiment-specific specialization limits transfer to other hands and prevents the reuse of demonstrations across heterogeneous hardware, leaving progress fragmented across isolated hand designs rather than advancing toward methods that generalize to new embodiments.

\noindent \textbf{Cross‑Embodiment Policy Learning.}
Recent work has increasingly explored how to share manipulation abilities across robotic hands with distinct morphologies~\cite{wei2024dro,li2022gendexgrasp,xu2025dexumi,he2025scaling,wu2025cedex,fei2025t,bauer2025latent,xu2024manifoundation, patel2025get}. Much of this progress has centered on grasping. One line of work focuses on intermediate grasp representations that abstract away embodiment-specific kinematics. These methods leverage representations such as interaction-centric fields, contact patterns, or PCA-based pose synergies to enable grasp transfer across different robotic hands, but remain largely confined to grasp synthesis and do not naturally extend to sequential manipulation skills~\cite{wei2024dro,li2022gendexgrasp,puang2025pchands}. Beyond grasping, some approaches target more general cross-embodiment behaviors through higher-level or embodiment-agnostic interfaces. Human-centric representations treat the human hand as a universal manipulation prior but typically assume human-like kinematics and require specialized hardware mappings~\cite{xu2025dexumi}. Particle-based dynamics learning provides an alternative by representing hands and objects as particle systems, yet is mainly applicable to structurally similar hands and constrained manipulation tasks~\cite{he2025scaling}. Despite these efforts, a unified cross-embodiment paradigm capable of supporting general manipulation across heterogeneous robotic hands is still lacking.

\section{Canonical Hand Representation Design}
\label{sec:design}

\noindent \textbf{Overview.} To enable learning policies that generalize across dexterous hands with different morphologies, we propose a parameterized canonical hand representation, which serves as the foundation for cross-embodiment manipulation. The goal is to express diverse robotic hands within a unified structural and kinematic framework that can be efficiently processed by learning-based models.

We begin by motivating the need for a canonical representation (Sec.~\ref{subsec:motivation}), highlighting the limitations of existing URDF formats and the challenges of defining a consistent and learnable description for heterogeneous dexterous hands. We then present the design of the canonical URDF (Sec.~\ref{subsec:urdf_design}), which captures shared human-inspired kinematic structure while enforcing consistent coordinate conventions. Next, we define the canonical parameter set that encodes key morphological and kinematic properties in a compact and interpretable form (Sec.~\ref{subsec:params}). We further describe the automatic parsing process that converts arbitrary hand URDFs into this canonical parameterization and generates standardized URDF models (Sec.~\ref{subsec:urdf_parse}). Finally, we establish a unified action space that aligns control dimensions across hands with different degrees of freedom (Sec.~\ref{subsec:action}), enabling a single policy to act consistently over diverse embodiments.

\begin{table}[t]
    \centering
    \caption{Comparison of representative dexterous robotic hands and their kinematic configurations. F, A, and R denote flexion, abduction/adduction, and axial rotation. DoF values in parentheses include passive or mechanically coupled joints.}
    \resizebox{\linewidth}{!}{
        \begin{tabular}{ccccccc}
        \toprule
        \multirow{2}{*}{\textbf{Name}} & 
        \multirow{2}{*}{\textbf{DoF}} & 
        \multicolumn{5}{c}{\textbf{Joint Types}}
        \\
        \cmidrule(lr){3-7}
        &
        &
        \textbf{Thumb} & 
        \textbf{Index} & 
        \textbf{Middle} & 
        \textbf{Ring} & 
        \textbf{Little}
        \\
        \midrule
        Shadow Hand & 22 & RAAFF & AFFF & AFFF & AFFF & FAFFF \\
        Sharpa Wave & 22 & AFAFF & AFFF & AFFF & AFFF & FAFFF \\
        WUJI Hand & 20 & FAFF & AFFF & AFFF & AFFF & AFFF \\
        DexHand 021 & 12(19) & AFFF & AFFF & FFF & AFFF & AFFF \\
        Orca Hand & 16 & FAFF & AFF & AFF & AFF & AFF \\
        Gaia Hand & 15 & AFF & AFF & AFF & AFF & AFF \\
        XHAND1 & 12 & AFF & AFF & FF & FF & FF \\
        Inspire & 6(12) & AFFF & FF & FF & FF & FF \\
        Allegro & 16 & ARFF & RFFF & RFFF & - & RFFF \\
        LEAP Hand & 16 & ARFF & AFFF & AFFF & - & AFFF \\
        Barrett & 8 & FF & AFF & AFF & - & - \\
        Dex3 & 7 & RFF & FF & FF & - & - \\
        \bottomrule
        \end{tabular}
    }
    \vspace{-10pt}
    \label{tab:dexhand}
\end{table}

\vspace{-2pt}
\subsection{Motivation}  \label{subsec:motivation}
\vspace{-1pt}
To develop a task-agnostic policy that generalizes across dexterous hands with diverse embodiments, the learning framework must incorporate a representation of the current hand as part of the model input. Equally important is a consistent action space across hands, enabling a single policy to operate seamlessly over different morphologies.

A natural approach is to describe each hand using general 3D representations, such as point clouds or signed distance fields. While these capture detailed geometry, they only reflect static structure and ignore the kinematic and dynamic properties crucial for manipulation. In contrast, the Unified Robot Description Format (URDF) encodes comprehensive information about a hand’s morphology and motion capabilities, including joint topology, link dimensions, and actuation limits. However, URDFs are typically handcrafted, vary across platforms, and are tree-structured and heterogeneous, making them difficult to use directly in learning-based pipelines.

To overcome these limitations, we introduce a canonical URDF representation that unifies the structural and kinematic description of dexterous hands. Each hand is represented as a compact set of parameters capturing key geometric and motion attributes, from which a standard URDF can be automatically generated. This canonical form preserves both structural and functional consistency while providing a format suitable for neural network input. It enables policies to be conditioned on hand morphology and executed in a shared action space, laying the foundation for scalable cross-hand generalization.

\subsection{Canonical URDF Design}  \label{subsec:urdf_design}
The canonical URDF unifies the structural representation of diverse dexterous hands by capturing shared kinematic principles in a standardized yet expressive format. The main challenge is selecting essential morphological parameters while remaining compatible with heterogeneous joint configurations and mechanical layouts. Preserving all native URDF details introduces numerous coordinate-dependent variables and inconsistent representations, while an overly compact model would constrain expressiveness. The canonical formulation therefore strikes a balance between representational richness, generality, and physical realizability.

\subsubsection{Kinematic Analysis of Existing Dexterous Hands}
To guide the canonical design, we analyzed a representative set of commercial and open-source dexterous hands, focusing on kinematic organization and degrees of freedom (DoF) (Table~\ref{tab:dexhand}). Despite mechanical differences, most designs share a human-inspired layout with recurring motion patterns. The thumb typically has two distal flexion joints and, when present, a separate abduction–adduction joint, while proximal thumb joints exhibit greater variability due to mechanical design. Remaining fingers generally have an abduction–adduction proximal joint followed by a flexion chain dominating grasping, with some higher-DoF hands adding extra flexion for the little finger to approximate human-like coupled motion.

\begin{figure}[t]
  \centering
  \begin{subfigure}{0.48\linewidth}
    \centering
    \includegraphics[width=0.55\linewidth]{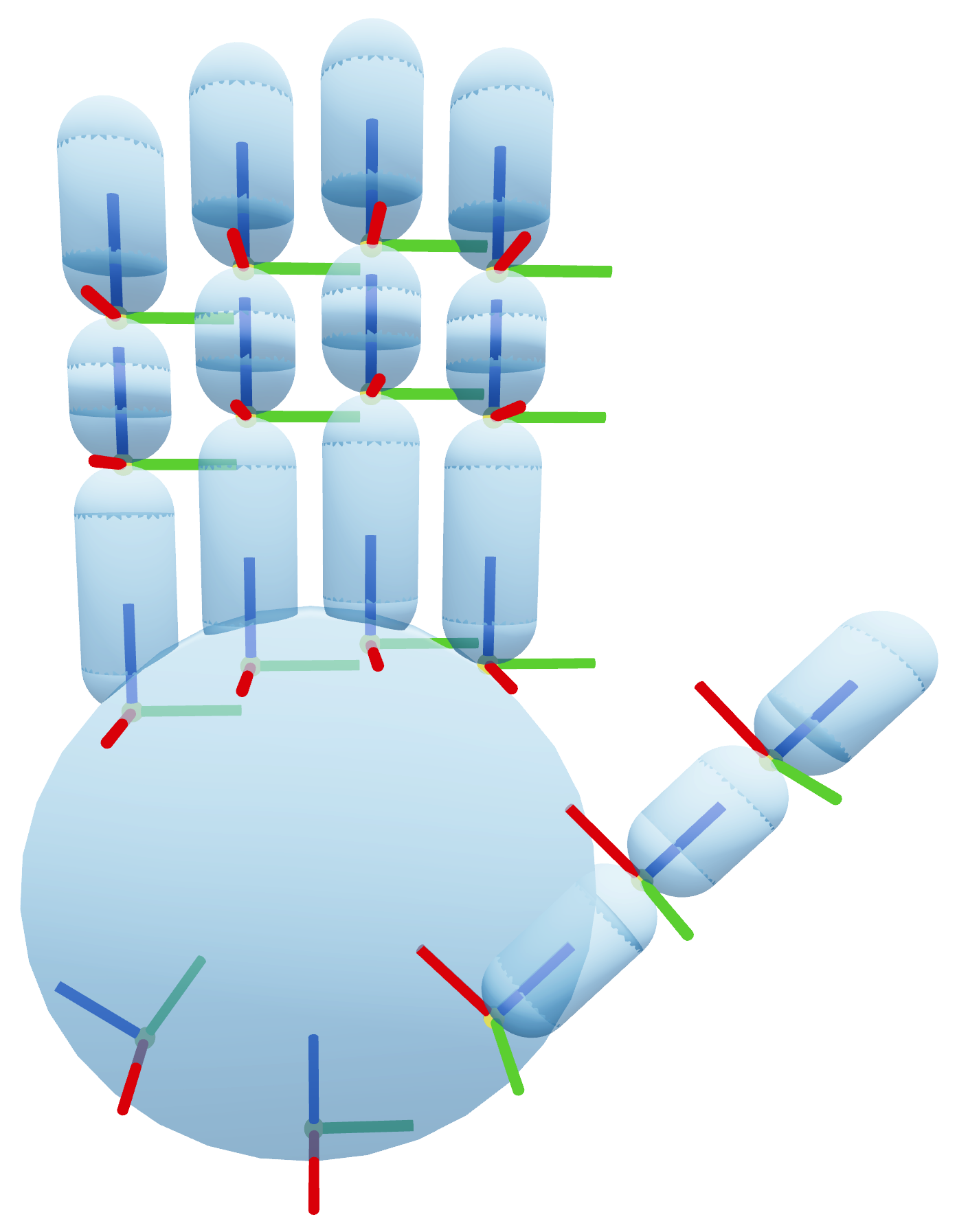}
    \captionsetup{skip=2pt}
    \caption{Mesh and frame visualization}
    \label{fig:urdf_vis}
  \end{subfigure}
  \begin{subfigure}{0.45\linewidth}
    \centering
    \includegraphics[width=0.55\linewidth]{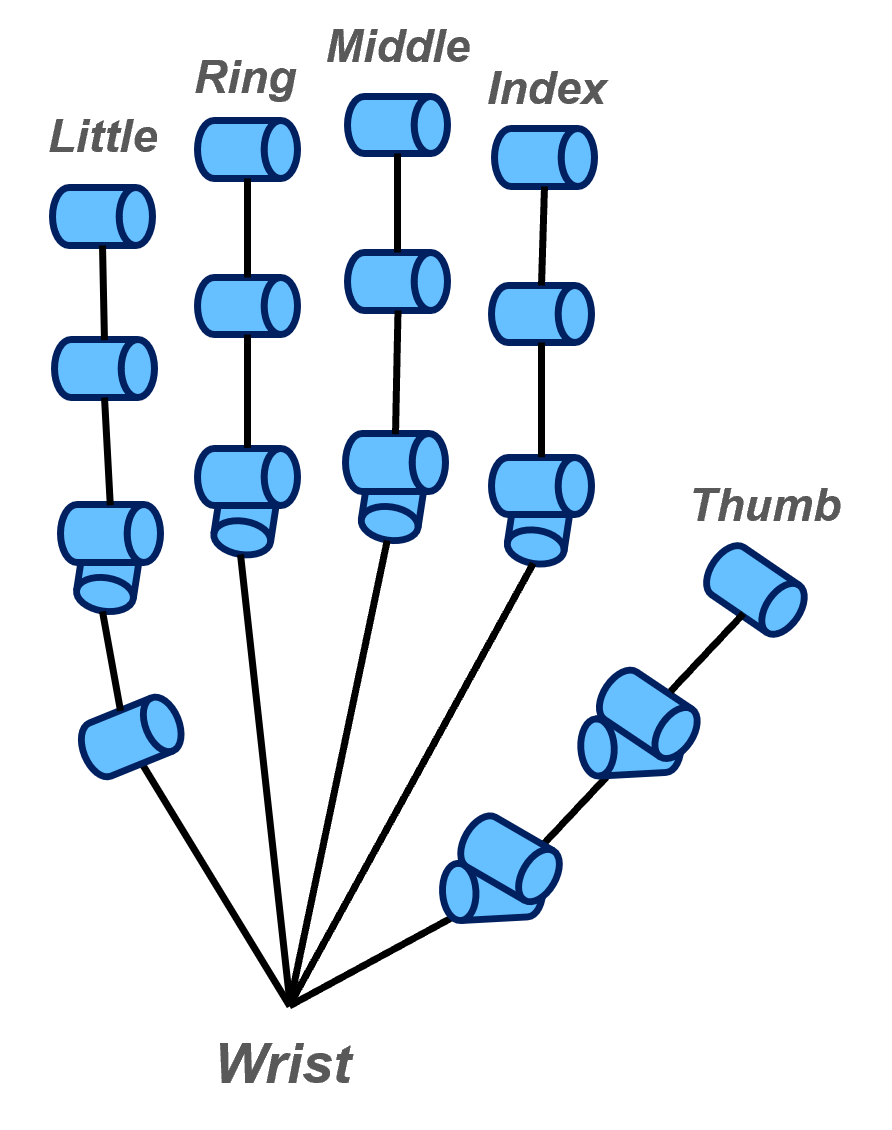}
    \captionsetup{skip=2pt}
    \caption{Kinetic skeleton diagram}
    \label{fig:urdf_kinetic}
  \end{subfigure}
  \caption{Structure of the canonical URDF. A right-hand configuration is shown for clarity, but the representation is applicable to both left- and right-handed hands.}
  \vspace{-15pt}
  \label{fig:canonical_urdf}
\end{figure}

Guided by these observations, we define a canonical URDF supporting up to five fingers and 22 DoF (Fig.~\ref{fig:canonical_urdf}), capturing the shared human-like topology. To maintain geometric consistency across different link meshes, all links are represented as capsule primitives, which also reflect the cylindrical shape of most hands’ collision meshes. This abstraction reduces geometric complexity while preserving the essential kinematic relationships required for dexterous manipulation.

\subsubsection{Unified Coordinate Conventions}
\begin{figure}[t]
  \centering
  \begin{subfigure}{0.45\linewidth}
    \hspace{15pt}
    \includegraphics[width=0.6\linewidth]{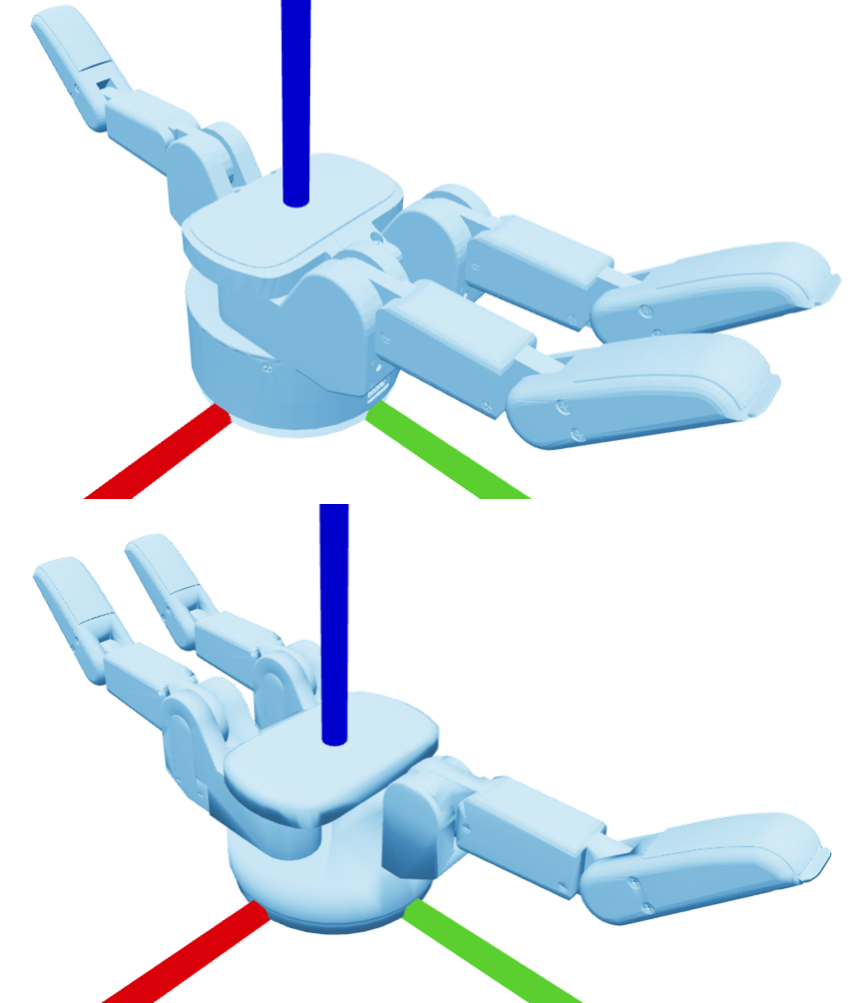}
    \vspace{-1pt}
    \caption{Global frame inconsistency}
    \label{fig:global}
  \end{subfigure}
  \begin{subfigure}{0.45\linewidth}
    \hspace{18pt}
    \includegraphics[width=0.6\linewidth]{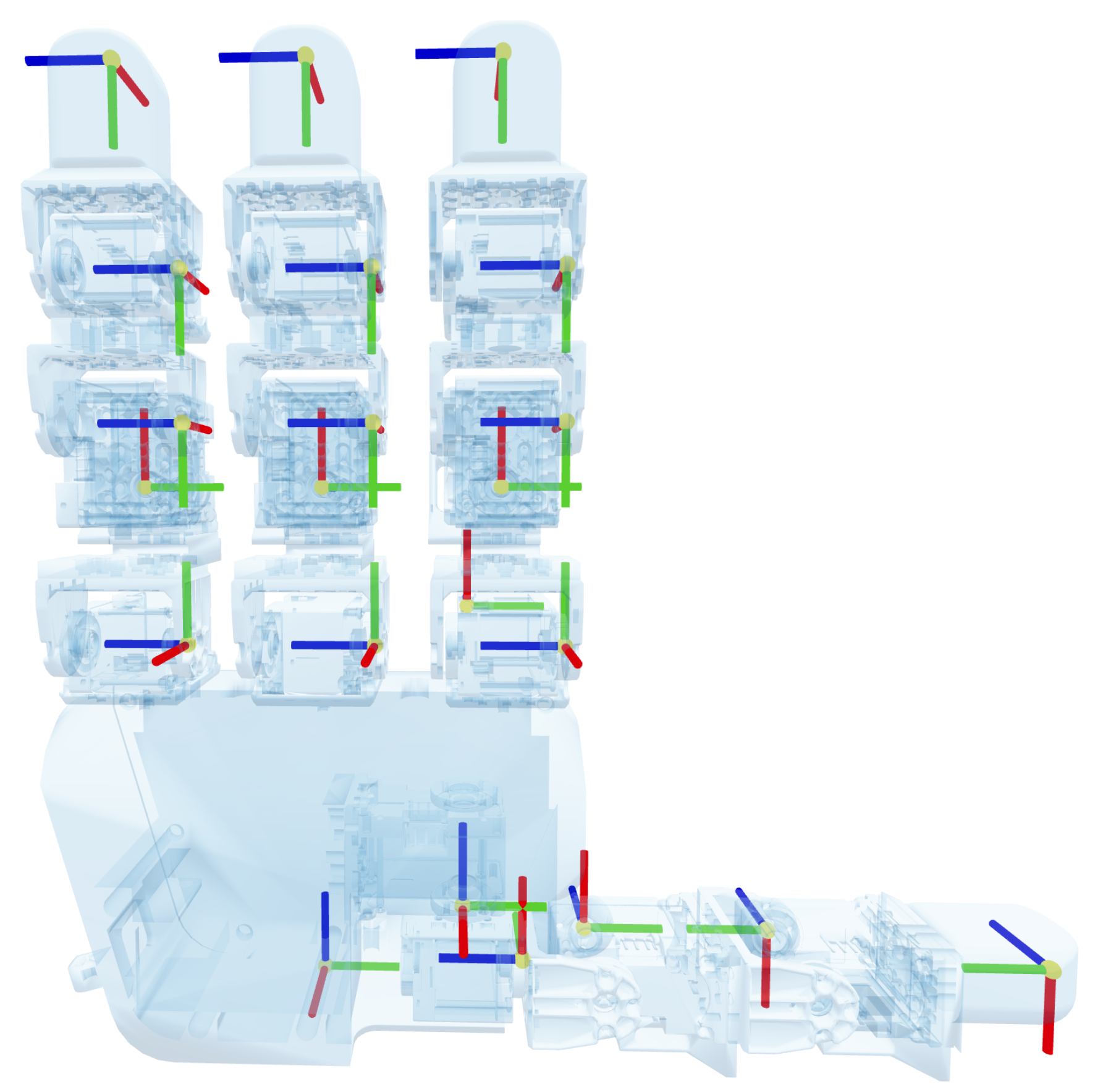}
    \vspace{-1pt}
    \caption{Local frame ambiguity}
    \label{fig:local}
  \end{subfigure}
  \caption{Coordinate frame inconsistencies in URDFs. (a) Global orientations vary across sources, (b) local joint frames use inconsistent axis definitions, leading to kinematic ambiguity.}
  \vspace{-12pt}
  \label{fig:frame}
\end{figure}

Dexterous hand URDFs vary in global and local coordinate frames. Even identical hardware can have different global axes (Fig.~\ref{fig:global}) or incompatible local joint orientations (Fig.~\ref{fig:local}), causing unnecessary transformations and ambiguity. Our canonical URDF enforces a unified convention aligned with human-hand kinematics (Fig.~\ref{fig:urdf_vis}): the palm normal along $+x$, the thumb (right hand) along $+y$, and other fingers along $+z$, with local axes $x$ for abduction–adduction, $y$ for flexion–extension, and $z$ for axial rotation when applicable. This standardized convention ensures consistent, interpretable kinematics and enables parameterization that captures key morphological and kinematic variations for cross-hand policy learning.

\subsection{Parameter Definition}  \label{subsec:params}
Building on the canonical URDF design, we define a compact parameter set for dexterous hands that captures key morphological and kinematic information. The canonical representation comprises 82 parameters, while a more comprehensive URDF template with 173 parameters is provided to support specialized hand designs. Full details and design rationale are given in Appendix~\ref{app:param}. 
By consolidating geometric and kinematic attributes into this structured representation, the canonical model captures the most salient variations across embodiments and facilitates efficient learning and transfer among diverse dexterous hands.

\subsection{URDF Parsing and Generation}  \label{subsec:urdf_parse}
To support a wide range of articulated hand designs, we develop an automatic framework for URDF parsing and generation. The framework extracts canonical parameters from arbitrary hand URDFs and reconstructs morphology and kinematics in the canonical space from minimal manual inputs, while also enabling full URDF generation from these parameters. Generation is implemented using the Jinja2 dynamic templating language~\cite{jinja}, allowing conditional inclusion of elements and automatic adaptation to hands with varying numbers of fingers or links. Together, these capabilities provide consistent bidirectional conversion between diverse robot models and the unified representation (Sec.~\ref{subsec:urdf_design}). Full details of the parsing and generation procedures are provided in Appendix~\ref{app:urdf}.

\vspace{-1pt}
\subsection{Action Space Unification}  \label{subsec:action}
\vspace{-1pt}
With the canonical URDF, all dexterous hands share a fixed 22-DoF structure defining a unified control and observation space. Hands with fewer active DoF have the corresponding joints set inactive and the associated links removed, maintaining a consistent joint and link representation across embodiments. Using the joint-to-joint mapping from parsing, we enable bidirectional conversion between original and canonical joint vectors with consistent indexing and sign conventions. This unified formulation standardizes joint dimensionality and ordering, providing a coherent action space for scalable cross-hand policy learning and transfer.

\vspace{-1pt}
\section{Canonical Representation Applications}
\vspace{-1pt}
\label{sec:application}

We evaluate the proposed canonical representation through four complementary studies. First, we examine whether the parameter space introduced in Sec.~\ref{subsec:params} induces a continuous and interpretable latent manifold (Sec.~\ref{subsec:latent}) using a variational autoencoder trained on diverse sampled hand morphologies. Second, we assess the physical fidelity of the canonical URDF by comparing in-hand object reorientation performance against the original hand models (Sec.~\ref{subsec:reori}), verifying that the canonical formulation preserves key kinematic and control properties. Third, leveraging the unified structure and action representation, we train a single grasping policy transferable across multiple dexterous hands (Sec.~\ref{subsec:grasp}), demonstrating scalability and cross-embodiment generalization. Finally, we investigate the zero-shot grasping capability enabled by hand conditioning using LEAP Hand and its variants (Sec.~\ref{subsec:zeroshot}), assessing generalization to unseen hand morphologies.

\begin{figure*}[t]
  \centering
  \includegraphics[width=\textwidth]{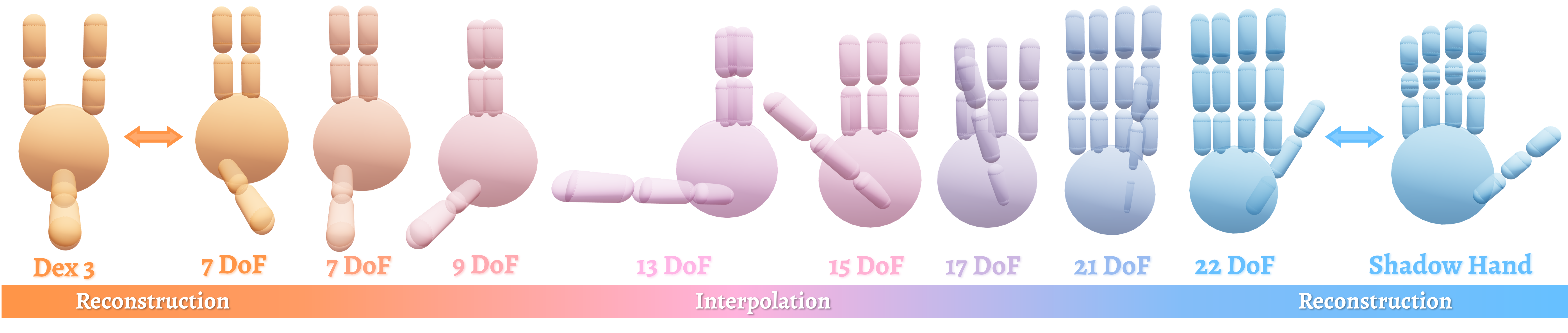}
  \vspace{-15pt}
  \caption{Visualization of latent-space interpolation between two dexterous hands. Canonical URDFs are shown at the ends, with decoded reconstructions and interpolated morphologies in between, demonstrating smooth transitions in DoF, finger arrangement, and overall geometry.}
  \vspace{-8pt}
  \label{fig:interp}
\end{figure*}

\subsection{Learning a Morphology Latent Space}  \label{subsec:latent}
To evaluate whether the canonical parameterization defines a compact and structured embedding, we train a variational autoencoder (VAE) to map hand morphology parameters into a 16-dimensional latent space. The latent representation captures essential geometric and kinematic variations across dexterous hands, forming a smooth manifold that supports interpolation and generalization.

A dataset of 65,536 synthetic hand configurations is generated by sampling each canonical parameter within physically feasible ranges. Continuous morphology parameters (palm/finger radii, link lengths) are sampled from bounded intervals, finger-origin translations preserve plausible proportions, joint axes are encoded as one-hot vectors over six canonical directions $(\pm x, \pm y, \pm z)$, and joint availability is encoded as binary indicators over the 22 canonical DoFs. The VAE reconstructs these parameters using type-specific losses with KL regularization. Full model architecture, training objectives, and hyperparameters are provided in Appendix~\ref{app:morp}.

\subsection{In-hand Reorientation}  \label{subsec:reori}

We evaluate the physical fidelity of the canonical representation using an in-hand object rotation task, where reinforcement learning agents are trained to rotate an object about a predefined axis. Separate policies are learned for the original dexterous hands and their canonical counterparts. Experiments are conducted in the IsaacGym simulator~\cite{makoviychuk2021isaac} using the Shadow Hand~\cite{shadowhand} and the LEAP Hand~\cite{shaw2023leap}, following prior in-hand rotation setups~\cite{qi2023hand, shaw2023leap}. Observations include joint positions and velocities, PD controller targets, and the object’s pose and velocities. The reward promotes high object angular velocity while penalizing object drift, excessive hand motion, and torque usage. Policies are trained with Proximal Policy Optimization (PPO)~\cite{schulman2017proximal} using MLP-based agents. Full training details are provided in Appendix~\ref{app:inhand}.

\begin{figure}[t]
    \centering
    \includegraphics[width=0.85 \linewidth]{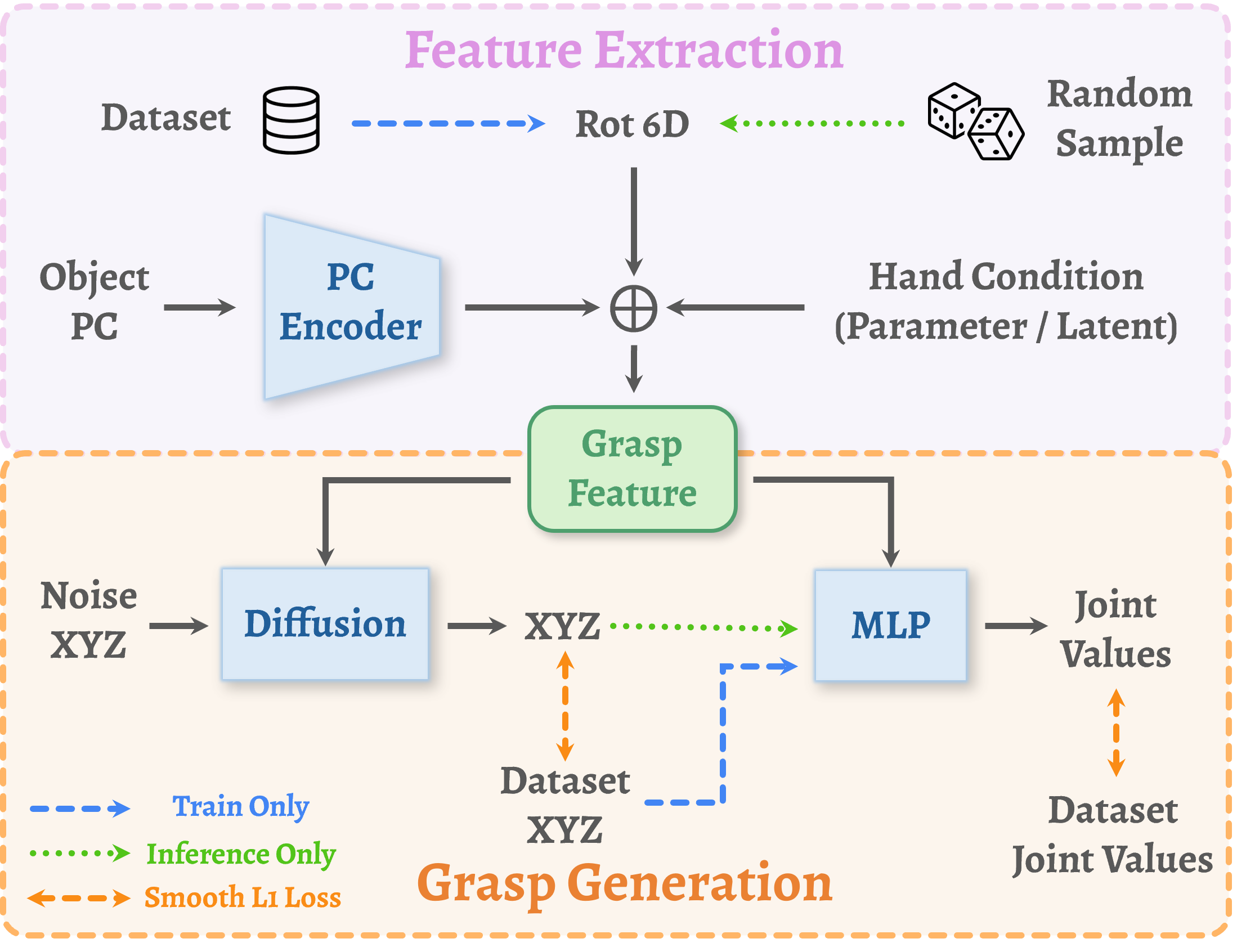}
    \vspace{-1pt}
    \caption{Two-stage cross-embodiment grasp generation pipeline.}
    \vspace{-15pt}
    \label{fig:grasp_pipeline}
\end{figure}

\subsection{Cross-Embodiment Dexterous Grasping}  \label{subsec:grasp}
To evaluate the transferability of control across different robotic hands, we train a cross-embodiment grasp pose prediction model that operates entirely within the canonical representation. A grasp pose is defined by the wrist pose $(T, R)$ relative to the object frame, parameterized by translation and rotation, together with the hand joint configuration $\theta$.

As shown in Fig.~\ref{fig:grasp_pipeline}, grasp generation follows a two-stage formulation inspired by~\citet{zhang2024dexgraspnet}. In the first stage, a diffusion-based conditional generator predicts the distribution of wrist translations around an object, conditioned on a grasp feature $f_g$ and an explicitly provided wrist rotation $R$. Providing $R$ separately decouples orientation from translation, allowing direct control over grasp direction and supporting diverse floating grasps. Training data spans multiple object orientations, enabling the model to generalize to arbitrary directions while supporting orientation-constrained sampling at test time. The second stage uses a lightweight MLP to predict the corresponding hand joint configuration $\theta$ conditioned on the sampled wrist pose $(T, R)$ and $f_g$, implementing a deterministic mapping.

The grasp feature $f_g$ integrates both object and hand morphology information. Object features are extracted from the input point cloud using the $\mathcal{D(R,O)}$ point-based encoder, while hand features are obtained from the frozen VAE latent embedding introduced in Sec.~\ref{subsec:latent}. This allows the model to generalize grasp predictions across hands with different morphologies.

The training objective combines diffusion-based translation prediction with deterministic joint regression, both optimized using a Smooth-$L_1$ loss:
\vspace{-3pt}
\begin{equation}
    \mathcal{L} = \text{SmoothL1}(\hat{T}, T) + \text{SmoothL1}(\hat{\theta}, \theta).
\vspace{-3pt}
\end{equation}

\subsection{LEAP Hand Zero-Shot Generalization} \label{subsec:zeroshot}
To further demonstrate the zero-shot generalization capability enabled by hand conditioning, and considering both the limited availability of existing dexterous hand designs and the feasibility of real-world experiments, we adopt the open-source, modular LEAP Hand hardware. By systematically varying the presence of individual links for each finger, we construct $4^4 = 256$ LEAP Hand variants, denoted as \texttt{leap\_xyzw}, where $x, y, z, w \in \{0, 1, 2, 3\}$ correspond to the number of links of the thumb, index, middle, and little fingers, respectively. For example, \texttt{leap\_3333} corresponds to the original LEAP Hand design.

Under our canonical hand representation, different LEAP Hand variants can be generated by simply modifying the corresponding morphology parameters, allowing efficient and scalable instantiation of a large number of hand designs. Grasp data for each variant are generated using Lightning Grasp~\cite{yin2025lightning}, where a dedicated configuration file is specified for each hand morphology. The generated grasps are further filtered using the data filtering strategy from $\mathcal{D(R,O)}$ Grasp~\cite{wei2024dro} to ensure physical plausibility. See Appendix~\ref{app:zeroshot} for more details.

To balance practical grasp feasibility and the evaluation of zero-shot generalization, we construct the grasping dataset using 66 LEAP Hand variants satisfying $x + y + z + w \geq 8$. The grasping policy is trained following the same procedure as in Sec.~\ref{subsec:grasp}, with the hand condition directly specified by the canonical morphology parameters. Zero-shot generalization is then evaluated on LEAP Hand variants with $x + y + z + w < 8$, as well as additional variants whose grasp data are entirely excluded during training, allowing us to assess generalization to unseen hand morphologies without further fine-tuning.

\section{Experiments}
\label{sec:exp}

We evaluate the effectiveness of the proposed canonical URDF representation and the applications described in Sec.~\ref{sec:application}. Our experiments focus on four key aspects: (1) the quality and continuity of the learned morphology latent space, (2) the fidelity and dynamics preserved when motions are expressed using the canonical hand model, (3) the generalization of a single unified grasping policy across dexterous hands with diverse kinematic structures, and (4) the zero-shot capability enabled by hand conditioning. We further deploy the zero-shot grasping policy on physical hardware, demonstrating that models trained in simulation with the canonical model maintain strong performance in real-world grasping tasks.

\subsection{VAE Latent Space Visualization}
We first examine whether the variational autoencoder (VAE) introduced in Sec.~\ref{subsec:latent} learns a structured, continuous embedding of dexterous-hand morphology. To assess this, we visualize latent-space interpolations between two hands with distinct geometric and kinematic designs, such as a compact three-finger gripper and a high-DoF anthropomorphic hand.

Given latent features $z_a$ and $z_b$ for the two canonical encodings, we interpolate using $z(\alpha) = (1-\alpha)z_a + \alpha z_b$ for $\alpha \in [0,1]$, decode each feature into canonical URDF parameters, and generate the corresponding hand models. As shown in Figure~\ref{fig:interp}, the interpolated structures evolve smoothly, with gradual variation in palm size, finger count and spacing, thumb placement, and DoF, indicating that the VAE captures a continuous morphological representation.

These results suggest that the latent space preserves intrinsic structural relationships among diverse hand designs, which is critical for downstream tasks such as morphology-conditioned grasping and cross-hand policy transfer.

\subsection{Canonical Hand Fidelity}

\subsubsection{In-Hand Reorientation}
We evaluate the physical fidelity of the canonical hand representation by comparing its performance to that of the original hand models in high-dynamic in-hand manipulation tasks, assessing whether essential kinematic and control properties are preserved.

\begin{figure}[t] \centering
    \includegraphics[width=0.8 \linewidth]{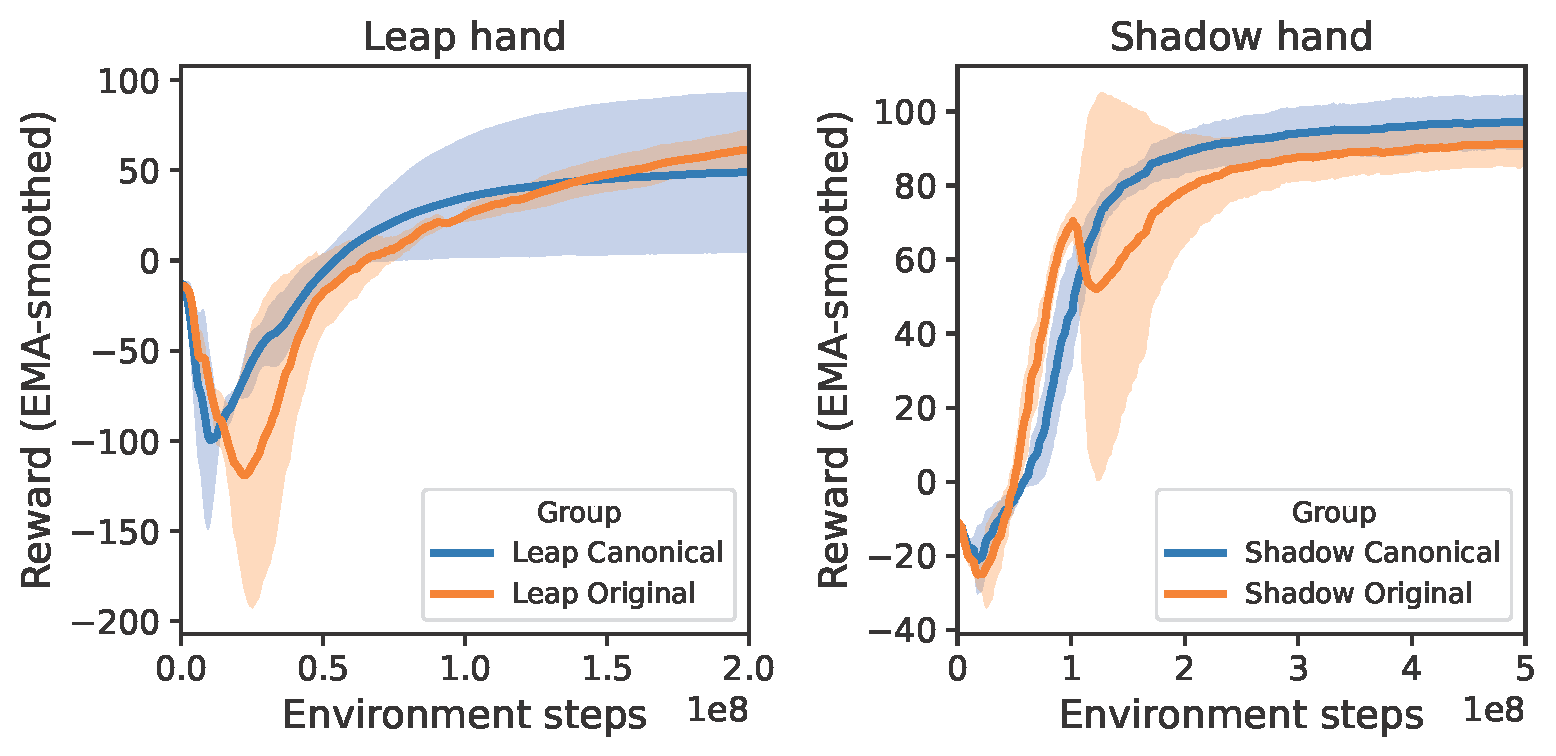}
    \caption{Smoothed log of training reward over simulation steps during training. The solid lines mark the mean value, and the shaded area consists of the standard deviation. }
    \vspace{-3pt}
    \label{fig:reward_curve}
\end{figure}

\begin{table}[t]
    \centering
    \caption{Comparison of in-hand reorientation policies trained with the canonical URDF vs. the original URDF.}
    \resizebox{\linewidth}{!}{
    \begin{tabular}{ccc}
        \toprule
        \textbf{Policy} & \textbf{Steps-to-Fall $\uparrow$} & \textbf{Cumulative Rotation $\uparrow$} \\
        \midrule
        Shadow (Original) & 369.66 & 9.09 \\
        Shadow (Canonical) & 390.62 & 10.92 \\
        \midrule
        LEAP (Original) & 397.62 & 5.63 \\
        LEAP (Canonical) & 326.98 & 6.31 \\
        \bottomrule
    \end{tabular}}
    \vspace{-10pt}
    \label{tab:inhand-rotation-results}
\end{table}

\begin{table}[t]
    \centering
    \caption{Comparison of grasp success rates when transferring across canonical and original URDFs.}
    \renewcommand\arraystretch{1.35}
    \vspace{-3pt}
    \resizebox{\linewidth}{!}{
        \begin{tabular}{c | >{\hspace{-10pt}} c >{\hspace{-25pt}} c >{\hspace{-10pt}} c}
            \toprule
            \addlinespace[-0.3px]
            \multirow{2}{*} {\textbf{Method}} 
            & \multicolumn{3}{c}{\textbf{Success Rate (\%)}}
            \\ 
            \cline{2-4} 
            & \textbf{Allegro} & \textbf{Barrett} & \textbf{ShadowHand}
            \\ \hline
            Ours (Canonical) & 84.20 & 88.10 & 62.90 
            \\
            Ours (Original) & \hspace{22.8pt} 71.60 \scriptsize{\color{colorblue}{(-12.60)}} & \hspace{21.2pt} 88.70 \scriptsize{\color{colorblue}{(+0.60)}} & \hspace{19.2pt} 62.60 \scriptsize{\color{colorblue}{(-0.30)}} 
            \\ \hline
            D(R,O) (Original) & 92.30 & 87.30 & 83.00 
            \\
            D(R,O) (Canonical) & \hspace{21pt} 92.38 \scriptsize{\color{colorblue}{(+0.08)}} & \hspace{21pt} 87.34 \scriptsize{\color{colorblue}{(+0.04)}} & \hspace{20pt} 78.63 \scriptsize{\color{colorblue}{(-4.37)}} 
            \\ [-1.6pt]
            \bottomrule
        \end{tabular}
    }
    \vspace{-6pt}
    \label{tab:transfer}
\end{table}

\noindent \textbf{Metrics.}
(1) \emph{Steps-to-Fall}: the number of simulation steps the object remains stably grasped; and (2) \emph{Cumulative Rotation}: the total rotation about the positive z-axis in radians, indicating how effectively the policy drives rotation. These metrics capture both stability and rotation, allowing direct comparison between canonical and original hands.

\noindent \textbf{Results.}
Table~\ref{tab:inhand-rotation-results} reports the average performance, showing that canonical representations achieve comparable Steps-to-Fall and Cumulative Rotation. Fig.~\ref{fig:reward_curve} further indicates that the learning progress and convergence patterns are similar for both representations. These results demonstrate that the canonical parameterization preserves essential manipulation dynamics and can serve as a practical interface for downstream applications such as reinforcement learning.

\subsubsection{Grasp Policy Transfer via Action Mapping}
Using the bidirectional mappings introduced in Sec.~\ref{subsec:action}, we replay grasp policies between the canonical and original URDF spaces. The policy trained in canonical space (Ours, Sec.~\ref{subsec:grasp_result}) is mapped to the original URDF, while the policy trained on the original URDF ($\mathcal{D(R,O)}$) is mapped to the canonical space. As shown in Table~\ref{tab:transfer}, transfers in both directions achieve closely matched success rates, indicating that the canonical representation preserves the action semantics required for stable execution. The main discrepancy occurs with the Allegro Hand, due to the omission of its axial-rotation joint in the canonical URDF, creating a minor structural mismatch. Overall, these results demonstrate that the mapping is robust and that the canonical action space provides a coherent interface for executing policies across diverse hand designs.

\begin{table}[t]
    \centering
    \vspace{5pt}
    \caption{Grasp performance comparison.}
    \renewcommand\arraystretch{1.2}
    \vspace{-3pt}
    \resizebox{\linewidth}{!}{
        \begin{tabular}{c|ccc|c}
            \toprule
            \addlinespace[-0.3px]
            \multirow{2}{*} {\textbf{Method}} 
            & \multicolumn{3}{c|}{\textbf{Success Rate (\%) $\uparrow$}}
            & \multirow{2}{*} {\textbf{Time (sec.) $\downarrow$}}
            \\ 
            \cline{2-4} 
            & \textbf{Allegro} & \textbf{Barrett} & \textbf{ShadowHand}
            \\ \hline
            DFC & 76.2 & 86.3 & 58.8 & $>$1800 
            \\
            GenDexGrasp & 51.0 & 67.0 & 54.2 & 19.71 
            \\
            D(R,O) Grasp & \textbf{92.3} & \underline{87.3} & \textbf{83.0} & 0.65 
            \\
            Ours & \underline{84.2} & \textbf{88.1} & \underline{62.9} & \textbf{0.13} 
            \\ [-1.6pt]
            \bottomrule
        \end{tabular}
    }
    \vspace{-12pt}
    \label{tab:success}
\end{table}

\subsection{Cross-Embodiment Grasping Performance} \label{subsec:grasp_result}
We build upon the filtered GenDexGrasp~\cite{li2022gendexgrasp} dataset provided by $\mathcal{D(R,O)}$ Grasp~\cite{wei2024dro}, which contains 24,764 valid grasps from three dexterous hands: Allegro, Barrett, and Shadow Hand. These hands vary substantially in geometry and kinematics, spanning 8–22 DoFs and three-, four-, and five-finger topologies. All grasps are converted into our canonical URDF, providing a consistent action space across embodiments and enabling evaluation of its generality and applicability across different hands.

We then train a single unified grasp-generation model on the canonical URDF introduced in Sec.~\ref{subsec:grasp}, evaluating its performance on 10 unseen objects against state-of-the-art baselines and comparing unified training with embodiment-specific training. The evaluation metric is provided in the Supplementary Material.

\begin{table}[t]
    \centering
    \caption{Comparison of grasp success rates. ``Specific'' indicates that each embodiment is trained independently, while ``Unified'' denotes joint training across all embodiments.}
    \renewcommand\arraystretch{1.2}
    \vspace{-1pt}
        \begin{tabular}{c|ccc}
            \toprule
            \addlinespace[-0.3px]
            \multirow{2}{*} {\textbf{Method}} 
            & \multicolumn{3}{c}{\textbf{Success Rate (\%)}}
            \\ 
            \cline{2-4} 
            & \textbf{Allegro} & \textbf{Barrett} & \textbf{ShadowHand}
            \\ \hline
            Specific & 82.1 & 87.6 & 55.4 
            \\
            Unified & \textbf{84.2} & \textbf{88.1} & \textbf{62.9} 
            \\ [-1.6pt]
            \bottomrule
        \end{tabular}
    \vspace{-15pt}
    \label{tab:unified}
\end{table}

\textbf{Comparison with State-of-the-Art Methods.}
As shown in Table~\ref{tab:success}, our lightweight model achieves grasp success rates comparable to more complex pipelines, without requiring optimization-based refinement. Inference uses a 10-step DDIM sampler~\cite{song2020denoising} and runs in only 0.13~s, making it the most efficient among the evaluated methods.

This comparison is intended to evaluate the canonical URDF as a downstream action space rather than to introduce a new grasping algorithm. The results show that even a simple model trained in this representation produces high-quality grasps across diverse hands, demonstrating that the canonical parameterization is expressive, coherent, and supports robust cross-hand grasp generation without reliance on hand-specific architectures or heavy engineering.

\textbf{Unified Training vs. Embodiment-Specific Training.}
The unified model consistently outperforms embodiment-specific models (Table~\ref{tab:unified}), indicating that the canonical URDF enables effective policy sharing across morphologies. By learning in a shared action space, hands with distinct kinematics benefit from each other’s data, highlighting the representation’s capacity to support cross-embodiment generalization.

\begin{table*}[htbp]
    \centering
    \caption{Real-World grasp success rates.}
    \renewcommand\arraystretch{1.2}
    \resizebox{0.9\textwidth}{!}{
        \begin{tabular}{c|cccccccccc|c}
            \toprule
            \addlinespace[-0.3px]
            \multirow{2}{*} {\textbf{Model}} 
            & \multicolumn{11}{c}{\textbf{Success Rate}}
            \\ 
            \cline{2-12} 
            & Apple & Band Aid & Coke & Cube & Football & Mayo & Orange & Pear & Sheep & Soccer & Average
            \\ \hline
            \texttt{leap\_3333} (trained)
            & 8/10 & 7/10 & 9/10 & 7/10 & 10/10 & 6/10 & 8/10 & 9/10 & 10/10 & 9/10 & 83/100
            \\ \hline
            \texttt{leap\_3033} (trained)
            & 8/10 & 8/10 & 2/10 & 6/10 & 9/10 & 6/10 & 7/10 & 9/10 & 10/10 & 10/10 & 75/100
            \\
            \texttt{leap\_3033} (zero-shot)
            & 8/10 & 10/10 & 5/10 & 5/10 & 7/10 & 2/10 & 9/10 & 7/10 & 9/10 & 9/10 & 71/100
            \\ \hline
            \texttt{leap\_3303} (trained)
            & 7/10 & 8/10 & 5/10 & 3/10 & 9/10 & 4/10 & 9/10 & 7/10 & 9/10 & 9/10 & 70/100
            \\
            \texttt{leap\_3303} (zero-shot)
            & 9/10 & 6/10 & 4/10 & 5/10 & 9/10 & 5/10 & 8/10 & 6/10 & 9/10 & 10/10 & 71/100
            \\ [-1.6pt]
            \bottomrule
        \end{tabular}
    }
    \label{tab:real}
    \vspace{-5pt}
\end{table*}

\begin{figure*}[t]
    \centering
    \vspace{5pt}
    \begin{subfigure}[b]{0.1\linewidth}
        \centering
        \includegraphics[width=\linewidth]{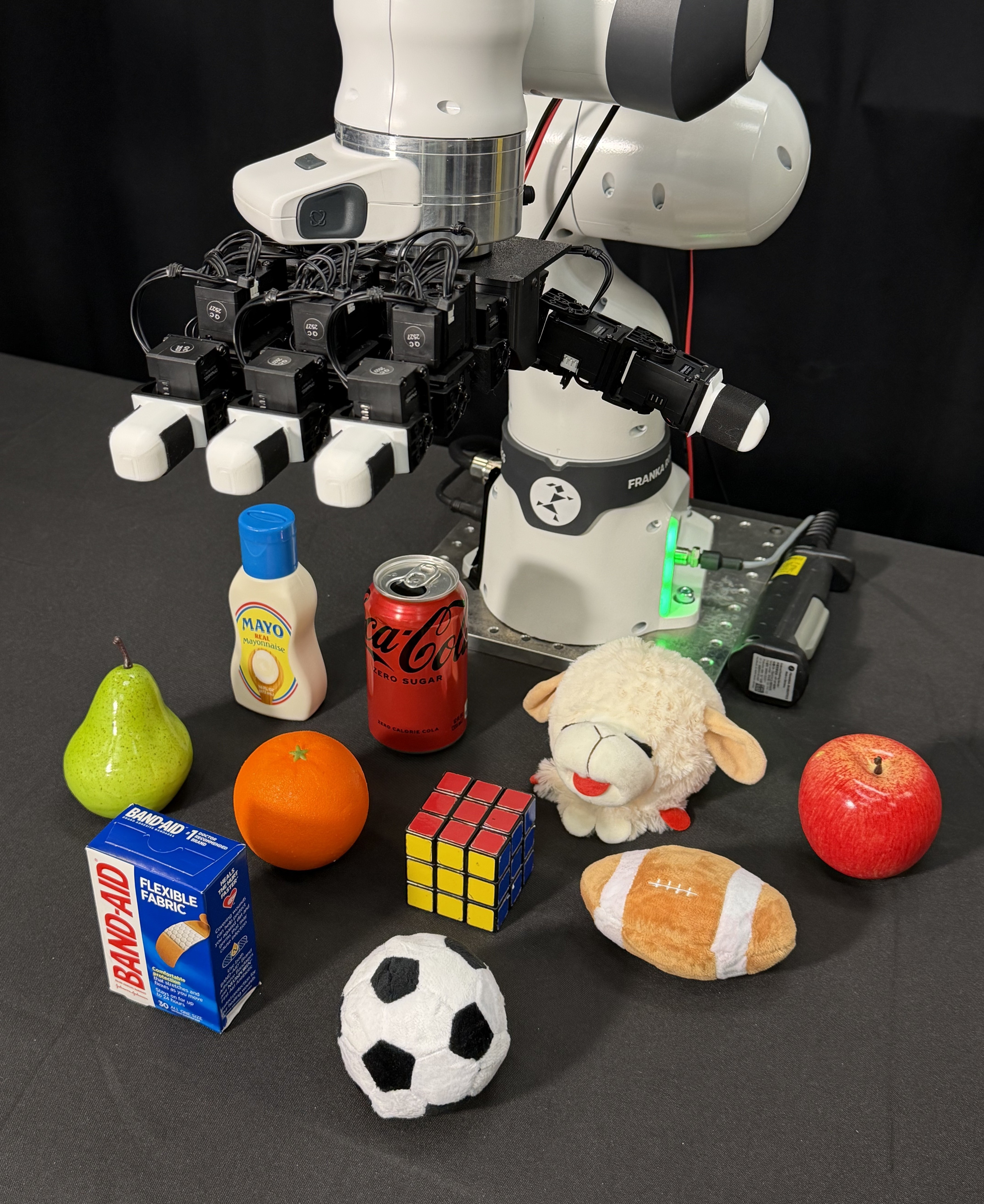}
        \caption{Object list}
        \label{fig:real_objects}
    \end{subfigure}
    \hfill
    \begin{subfigure}[b]{0.88\linewidth}
        \centering
        \includegraphics[width=\linewidth]{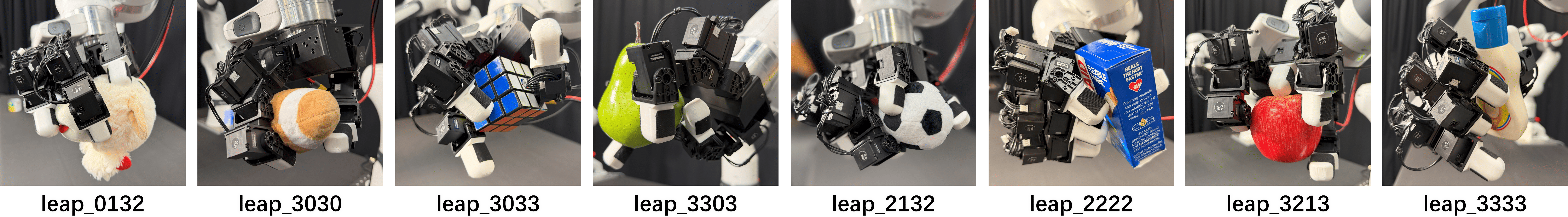}
        \caption{Zero-shot (left four) and trained (right four) grasping results with different LEAP Hand variants.}
        \label{fig:real_hands}
    \end{subfigure}
    \caption{Real-world grasping objects and results.}
    \vspace{-5pt}
    \label{fig:real}
\end{figure*}

\begin{table}[t]
    \centering
    \caption{Comparison of grasp success rates. Underlined values indicate zero-shot evaluation.}
    \renewcommand\arraystretch{1.2}
    \resizebox{\linewidth}{!}{
        \begin{tabular}{c|ccc}
            \toprule
            \addlinespace[-0.3px]
            \multirow{2}{*} {\textbf{Model}} 
            & \multicolumn{3}{c}{\textbf{Success Rate (\%)}}
            \\ 
            \cline{2-4} 
            & \textbf{\texttt{leap\_3033}} & \textbf{\texttt{leap\_3303}} & \textbf{\texttt{leap\_3330}}
            \\ \hline
            All Data & 76.1 & \textbf{85.4} & 43.3 
            \\ \hline
            No \texttt{leap\_3033} Data & \underline{67.8} & 83.4 & 31.5 
            \\
            No \texttt{leap\_3303} Data & \textbf{81.5} & \underline{81.9} & \textbf{46.9} 
            \\
            No \texttt{leap\_3330} Data & 74.7 & 81.6 & \underline{36.3} 
            \\ [-1.6pt]
            \bottomrule
        \end{tabular}
    }
    \vspace{-3pt}
    \label{tab:zero-shot_1}
\end{table}

\begin{figure}[t]
\begin{minipage}[t]{0.54\linewidth}
    \centering
    \captionsetup{width=0.9\linewidth}
    \captionof{table}{Grasp success rates and dataset number for selected LEAP Hand variants.}
    \renewcommand\arraystretch{1.39}
    \setlength{\tabcolsep}{3pt}
    \resizebox{1\linewidth}{!}{%
        \begin{tabular}{c|ccccc}
            \toprule
            \addlinespace[-0.3px]
            \multirow{2}{*} {\textbf{Model}} 
            & \multicolumn{5}{c}{\textbf{Success Rate (\%)}}
            \\ 
            \cline{2-6} 
            & \textbf{\texttt{0303}} & \textbf{\texttt{0312}} & \textbf{\texttt{2203}} & \textbf{\texttt{3030}} & \textbf{\texttt{3103}}
            \\ \hline
            All Data & 46.9 & \textbf{13.3} & \textbf{65.1} & 36.2 & \textbf{46.6} 
            \\
            Specific Data & \textbf{75.1} & 12.1 & 33.9 & \textbf{55.4} & 18.5 
            \\ \hline
            Data Num & 37249 & 4368 & 2458 & 37217 & 2124 
            \\ [-1.6pt]
            \bottomrule
        \end{tabular}
    }
    \vspace{-8pt}
    \label{tab:zero-shot_2}
\end{minipage}
\hfill
\begin{minipage}[t]{0.45\linewidth}
    \centering
    \captionsetup{width=0.9\linewidth}
    \captionof{table}{Grasp results for \texttt{leap\_3033} across hand conditions.}
    \setlength{\tabcolsep}{2pt}
    \renewcommand\arraystretch{1.15}
    \resizebox{\linewidth}{!}{
        \begin{tabular}{c|cc}
            \toprule
            \addlinespace[-0.3px]
            \multirow{2}{*} {\textbf{Condition}} 
            & \multicolumn{2}{c}{\textbf{Success Rate (\%)}}
            \\ 
            \cline{2-3} & \textbf{All Data} & \textbf{Zero-Shot}
            \\ \hline
            \texttt{leap\_3303} & \textbf{85.4} & \textbf{81.6}
            \\ \hline
            \texttt{leap\_3033} & 33.9 & 12.8 
            \\
            \texttt{leap\_3330} & 20.5 & 2.4 
            \\
            \texttt{leap\_3333} & 85.1 & 71.9 
            \\ [-1.6pt]
            \bottomrule
        \end{tabular}
    }
    \vspace{-12pt}
    \label{tab:wrong_condition}
\end{minipage}
\end{figure}

\subsection{Zero-Shot Grasping Performance}
We evaluate the zero-shot generalization of the canonical grasping policy using a set of 10 objects. Following the procedure described in Sec.~\ref{subsec:zeroshot}, we generate and filter grasps for various LEAP Hand variants, limiting each hand–object pair to a maximum of 200 grasps and applying selection criteria on the hand variants. The resulting dataset contains 69,917 grasps. For zero-shot evaluation, the training data excludes grasps corresponding to the validated LEAP Hand variants.

We first train a full model on the complete dataset to assess cross-embodiment performance on seen hands. We then train three additional models, each excluding data from a specific LEAP Hand variant (\texttt{leap\_3033}, \texttt{leap\_3303}, and \texttt{leap\_3330}) for zero-shot testing. As shown in Table~\ref{tab:zero-shot_1}, the models conditioned on hand morphology achieve performance on unseen hands comparable to that on seen hands, demonstrating strong zero-shot transfer capability across unobserved hand designs.

We further evaluate zero-shot generalization on a set of LEAP Hand variants with $x+y+z+w < 8$, which differ substantially from those in the training dataset, making the task more challenging. To isolate the effect of hand morphology on success rates, we generated and filtered grasp data specifically for these variants, training models only on their corresponding data. The resulting grasp data numbers and success rates are summarized in Table~\ref{tab:zero-shot_2}. 

As shown, the zero-shot models outperform the variant-specific models on \texttt{leap\_0312}, \texttt{leap\_2203}, and \texttt{leap\_3103}. Performance is slightly lower on \texttt{leap\_0303} and \texttt{leap\_3030}, likely because these two variants have only two fingers, resembling a gripper, and their grasp patterns differ markedly from those of dexterous hands in the training set. In addition, these variants generated a large number of grasps ($\approx$30k) in a single round, which benefits the variant-specific models. Nevertheless, the zero-shot models still demonstrate strong generalization to these challenging morphologies, highlighting the effectiveness of the canonical hand conditioning and unified action space.

To further validate the impact of hand conditioning, we evaluated the model using incorrect hand conditions on other LEAP Hand variants. As shown in Table~\ref{tab:wrong_condition}, we tested the variant \texttt{leap\_3033}. Overall, applying an incorrect hand condition substantially reduces grasp success rates. Using the original LEAP Hand parameters (\texttt{leap\_3333}) in the ``All Data'' trained model results in only a minor drop, as the model can partially overfit to the included variant. In contrast, in the zero-shot setting, success rates drop significantly, highlighting the critical role of hand conditioning.

\begin{figure}[t]
    \centering
    \begin{subfigure}{0.49\linewidth}
        \centering
        \includegraphics[width=\linewidth]{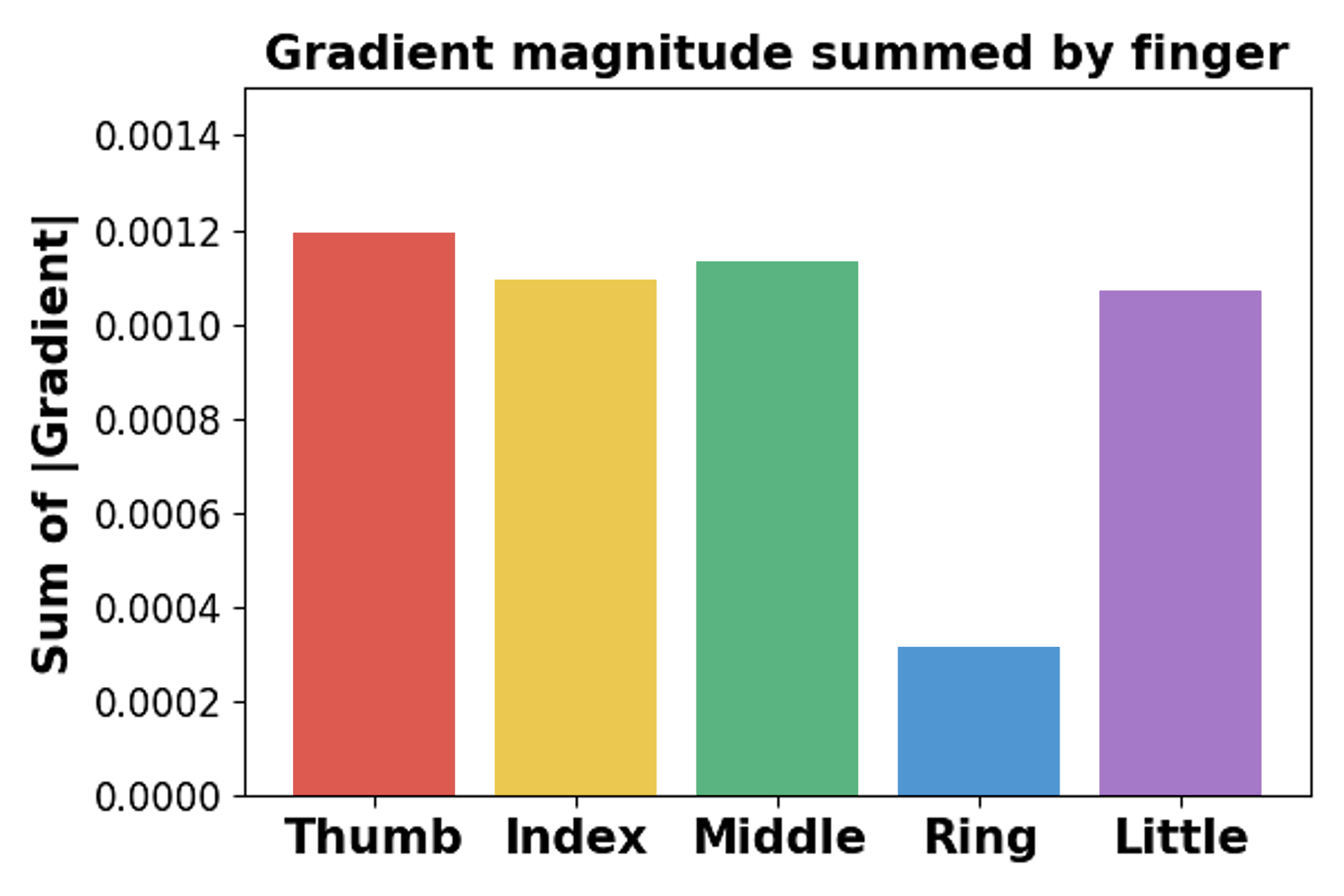}
        \captionsetup{skip=2pt}
        \caption{\texttt{leap\_3333}}
    \end{subfigure}
    \hfill
    \begin{subfigure}{0.49\linewidth}
        \centering
        \includegraphics[width=\linewidth]{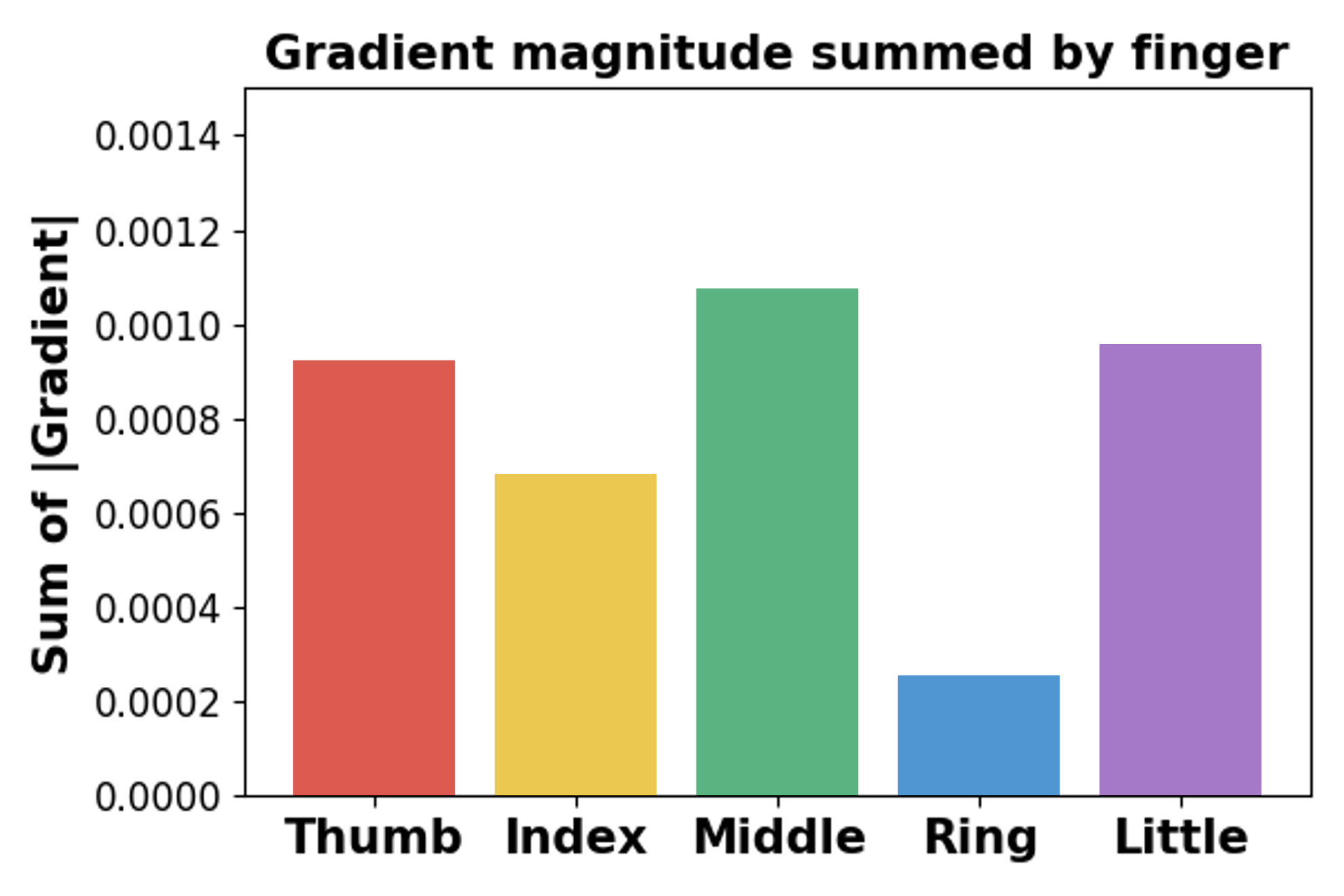}
        \captionsetup{skip=2pt}
        \caption{\texttt{leap\_3033}}
    \end{subfigure}
    \vspace{-10pt}
    \caption{Gradient magnitude visualization.}
    \vspace{-18pt}
    \label{fig:grad}
\end{figure}

We also visualize the backward gradients with respect to the canonical parameters (Fig.~\ref{fig:grad}). The gradient for the ring finger remains consistently low, as this finger is absent in the LEAP Hand. In variant \texttt{leap\_3033}, where the index finger is removed, its gradient drops markedly, indicating that the model has learned to focus on the functional fingers essential for successful grasps.

\subsection{Real-World Experiment}
To validate the sim-to-real transfer and practical applicability of our approach, we deploy the LEAP Hand~\cite{shaw2023leap} grasping policies on a Franka Research 3 robotic arm. Object observations are captured using an Intel RealSense L515 depth camera.  The evaluation is conducted on a set of 10 diverse objects, as illustrated in Fig.~\ref{fig:real_objects}, and across several LEAP Hand variants (Fig.~\ref{fig:real_hands}). Details of real-world experiments are in the Supplementary Material.

We evaluate both models trained on the canonical dataset and zero-shot models that have never seen the target hand variants. As summarized in Table~\ref{tab:real}, the trained models achieve high grasp success rates, demonstrating that the canonical hand representation preserves the essential dynamics and physical fidelity of the original hands, and that sim-to-real transfer is reliable. Importantly, the zero-shot models achieve success rates close to those of the trained models, highlighting strong generalization capabilities and the effectiveness of the hand condition in guiding grasping across unseen morphologies. The multi-panel illustrations in Fig.~\ref{fig:real_hands} further demonstrate that even for diverse and unusual LEAP Hand variants, the policy executes stable and robust grasps, highlighting the strength and effectiveness of the canonical hand conditioning.

\section{Conclusion and Discussion}
\label{sec:conclusion}

We introduced a canonical representation that maps heterogeneous dexterous hands into a shared parameter and action space, enabling scalable cross-embodiment learning. Its continuous morphology parameterization enables explicit hand-aware conditioning, and the unified action space facilitates data and policy sharing across platforms. Together, these properties enable embodiment-agnostic manipulation policies that generalize in a zero-shot manner to unseen hands, establishing the canonical URDF as a practical foundation for scalable cross-embodiment manipulation. Beyond dexterous hand grasping, our insights can extend to broader robotic embodiments such as humanoid robots and to more diverse manipulation tasks, potentially benefiting the embodied AI and robotics communities.

\bibliographystyle{plainnat}
\bibliography{references}

\clearpage
\appendix
\subsection{Canonical URDF Parameter Definition}  \label{app:param}

\subsubsection{Base Canonical Parameterization}
As introduced in Sec.~\ref{subsec:params}, we adopt a canonical parameterization of dexterous hands that abstracts a wide range of robotic hand designs into a compact and interpretable representation. Derived from a canonical URDF design, this parameterization captures the most salient morphological and kinematic variations across embodiments while removing redundant or simulator-specific details present in the original URDF files. In total, the canonical representation comprises 82 parameters, providing a unified description that can be directly used by learning-based models and supports cross-embodiment transfer.

\textbf{Morphological Parameters.}
The first component of the parameter set describes the global geometry of the hand. The palm is modeled as a cylindrical primitive, and each finger link is modeled as a capsule. Hand morphology is parameterized by \texttt{palm\_radius}, \texttt{finger\_radius}, and \texttt{finger\_lengths}.

To reduce dimensionality while preserving realistic proportions, we adopt two mild structural assumptions. First, all fingers share the same diameter, represented by a single \texttt{finger\_radius} parameter. Second, palm thickness is assumed to scale proportionally with the finger radius and therefore does not require an independent parameter. These assumptions are consistent with common robotic hand designs and introduce minimal loss of generality.

Finger link lengths are encoded using a small set of shared parameters. Specifically, three parameters represent the flexion chain shared by all non-thumb fingers, reflecting the empirical regularity that the index, middle, ring, and little fingers exhibit highly correlated phalangeal proportions in both human anatomy and robotic implementations. The thumb, whose morphology differs more substantially across designs, is encoded using three additional length parameters. As a result, \texttt{finger\_lengths} forms a six-dimensional vector in total.

This formulation preserves essential anthropomorphic structure while keeping the morphological parameter space compact and well-suited for learning.

\textbf{Kinematic Parameters.}
The second component of the canonical representation defines the kinematic structure of the hand, including joint placement, joint axes, and joint limits.

Finger base locations on the palm are encoded by translational offsets \texttt{finger\_xyz}. Based on the observation that non-thumb fingers are typically mounted perpendicular to the palm and approximately coplanar on the palm’s $yz$-plane, we model only their translational origins and omit rotational offsets. This simplification substantially reduces the parameter count while remaining faithful to common mechanical layouts.

The thumb requires additional flexibility due to its diverse mounting configurations and more complex kinematic role. We therefore include an explicit orientation parameter \texttt{thumb\_rpy}, which specifies the rotation of the thumb base relative to the palm frame. Following the kinematic analysis in Sec.~\ref{subsec:urdf_design}, only the proximal two thumb joints exhibit meaningful variability in joint axes across different hands. These are represented using \texttt{thumb\_axes}, while all remaining joint axes are fixed to canonical directions.

Finally, joint feasibility is specified by \texttt{joint\_lowers} and \texttt{joint\_uppers}, which define the allowable motion ranges for each of the 22 canonical degrees of freedom. Together, these parameters capture the essential motion characteristics of dexterous hands while abstracting away unnecessary implementation details.

\textbf{Modeling Assumptions.}
For clarity, we summarize the structural assumptions underlying the canonical parameterization. Joint indices $i$ increase from the finger base toward the fingertip:

\begin{enumerate}
    \item All fingers share the same capsule diameter; palm thickness is implicitly tied to this diameter.
    \item All non-thumb fingers use identical link lengths.
    \item All non-thumb fingers lie on the palm-aligned $yz$-plane.
    \item For all fingers, joint1 and joint2 share the same origin.
    \item For the thumb, joint3 and joint4 share the same origin.
    \item Except for the thumb’s joint1, all remaining joints have fixed local-frame orientations with $rpy=(0,0,0)$.
    \item All joint axes are fixed except for the thumb’s joint1 and joint2. Specifically, the thumb’s joint3 and all non-thumb joint1 use the $+x$ axis, while all remaining joints use the $+y$ axis.
\end{enumerate}

These assumptions capture the dominant structural regularities of dexterous hands while enabling a compact, standardized, and learning-friendly representation.

\begin{table}[t]
    \centering
    \setlength{\tabcolsep}{9.5pt}
    \caption{Canonical URDF parameter definition.}
    \vspace{-3pt}
    \resizebox{0.95\linewidth}{!}{
        \begin{tabular}{c c c}
        \toprule
        \textbf{Param Name} & 
        \textbf{Num} & 
        \textbf{Param Meaning}
        \\
        \midrule
        palm\_radius & 1 & radius of palm cylinder
        \\ [3pt]
        finger\_radius & 1 & radius of finger capsule
        \\ [3pt] 
        finger\_lengths & 6 & \makecell[c]{thumb link lengths and \\ finger link lengths}
        \\ [10pt] 
        finger\_xyz & 15 & knuckle origin translations of 5 fingers
        \\ [3pt]
        little\_extra\_origin & 6 & \makecell[c]{joint origin (xyz+rpy) of little \\ finger extra origin}
        \\ [10pt]
        thumb\_rpy & 3 & knuckle origin rotation of thumb
        \\ [3pt]
        thumb\_axes & 6 & thumb axes of proximal two joints
        \\ [3pt]
        joint\_lowers & 22 & lower ranges of all joints
        \\ [3pt]
        joint\_uppers & 22 & upper ranges of all joints
        \\ 
        \midrule
        \textbf{Total} & \textbf{82} & \\
        \bottomrule
        \end{tabular}
    }
    \vspace{-2pt}
    \label{tab:params}
\end{table}

\begin{table}[t]
    \centering
    \setlength{\tabcolsep}{9.5pt}
    \caption{Extended canonical URDF parameter definition.}
    \vspace{-3pt}
        \begin{tabular}{c c c}
        \toprule
        \textbf{Param Name} & 
        \textbf{Num} & 
        \textbf{Param Meaning}
        \\
        \midrule
        palm\_radius & 1 & radius of palm cylinder
        \\ [3pt]
        finger\_radii & 5 & radii of 5 fingers
        \\ [3pt] 
        finger\_lengths & 15 & link lengths of 5 fingers
        \\ [3pt] 
        joint\_origins & 72 & origins of 12 joints
        \\ [3pt]
        joint\_axes & 36 & axes of 12 joints
        \\ [3pt]
        joint\_lowers & 22 & lower ranges of all joints
        \\ [3pt]
        joint\_uppers & 22 & upper ranges of all joints
        \\ 
        \midrule
        \textbf{Total} & \textbf{173} & \\
        \bottomrule
        \end{tabular}
    \vspace{-10pt}
    \label{tab:params_extend}
\end{table}

\subsubsection{Extended Canonical Parameterization}  \label{app:extend}
While the base canonical parameterization is sufficiently expressive to model the vast majority of dexterous hand designs, certain embodiments exhibit structural deviations that cannot be captured exactly under the canonical assumptions. For example, in the Allegro Hand, the rotation axis of the non-thumb joint1 is aligned with the $+z$ direction rather than the canonical choice. In the LEAP Hand, the proximal two joints of each non-thumb finger are swapped in the kinematic tree, with the flexion joint located closer to the palm than the abduction/adduction joint. Although such designs can be approximated through reasonable canonical mappings, these cases may introduce small geometric or kinematic discrepancies.

To address these limitations, we additionally provide an extended parameterization that relaxes many of the canonical assumptions. As summarized in Table~\ref{tab:params_extend}, this representation contains 173 parameters and can exactly encode all observed hand designs with substantially reduced approximation error. This extended formulation demonstrates the extensibility of our framework and allows the parameter set to be expanded when higher fidelity is required or when modeling unconventional embodiments.

Compared with the base canonical design, the extended parameterization increases both geometric and kinematic expressiveness. On the morphology side, it replaces the shared \texttt{finger\_radius} and \texttt{finger\_lengths} with per-finger specifications, \texttt{finger\_radii} and expanded \texttt{finger\_lengths}, enabling each finger to have its own radius and link-length configuration. This allows the representation to capture finer geometric variation and to accommodate future dexterous hand designs with more diverse proportions.

On the kinematic side, the extended representation introduces additional joint origins and rotation axes for twelve joints. These include the first three joints of the thumb and the little finger, as well as the first two joints of the remaining three fingers. This expansion enables a broader range of mounting configurations and kinematic couplings to be represented explicitly. Under the extended parameter set, the only remaining assumptions are that the two distal joints of each finger share fixed local-frame orientations ($rpy = (0,0,0)$) and use the flexion-aligned $+y$ axis. These minimal constraints preserve a consistent convention for distal articulation while maximizing compatibility with structurally diverse dexterous hands.

\subsection{URDF Parsing and Generation Details}  \label{app:urdf}

This section describes the automatic framework used for URDF parsing and generation, which enables bidirectional conversion between the original and the canonical robot URDFs. We detail both the parameter extraction procedure and the canonical URDF generation process.

\subsubsection{Canonical Parameter Extraction}
To obtain canonical parameters from an original robotic hand design, we parse the corresponding URDF and extract the geometric and kinematic information required by our representation. The parser requires only two minimal inputs: (i) a mapping between original URDF joints and their canonical counterparts, and (ii) the canonical palm root transform in the world frame. All remaining quantities are inferred automatically. Below, we summarize the extraction procedure for each parameter group.

\textbf{Palm Geometry.}
The palm link is identified as the unique link that serves as the parent of multiple revolute joints. Its bounding box is computed from the associated mesh geometry, and the palm radius is estimated from the average in-plane dimensions of this bounding box. This provides a stable approximation of palm thickness and overall scale.

\textbf{Finger Geometry.}
Finger radii are estimated by examining all links belonging to finger kinematic chains. For each link, we compute the minimum dimension of its mesh bounding box, and these values are averaged across all fingers to obtain a consistent capsule radius. Finger link lengths are derived from joint-to-joint distances along each kinematic chain. The translational components of joint origins specify the first two segment lengths, while the third link length is approximated by averaging adjacent segments when an explicit fingertip frame is not available.

\textbf{Finger Base Positions.}
Finger base locations are computed by transforming the child link frame of each finger’s base joint into the canonical palm coordinate system. The palm’s pose is defined by \texttt{palm\_origin} metadata, enabling consistent conversion from world coordinates to the palm frame.

\textbf{Thumb Base Orientation.}
To determine \texttt{thumb\_rpy}, we use the frames of the thumb’s first and last existing joints. The vector from the thumb base to the thumb tip defines the local $+z$ direction, while the joint axis at the base provides the local $+y$ direction. Together, these constraints define a right-handed thumb coordinate frame, which is expressed in the palm frame and converted to Euler angles.

\textbf{Thumb Joint Axes.}
The rotation axes of the first two thumb joints are read directly from the URDF joint specifications. To express them in the canonical thumb frame, we compute the relative rotation between the original URDF joint frames and the reordered thumb base frame.

\textbf{Extra Little-Finger Orientation.}
For hand designs that include an additional abduction joint at the base of the little finger, we extract its joint frame from the URDF and reorder its axes using the same convention as the thumb (palm-up direction as $+x$, rotation axis as $+y$). This yields the \texttt{little\_extra\_origin} parameters in the extended representation.

\textbf{Joint Limits.}
Joint lower and upper bounds are read directly from the URDF \texttt{limit} tags for each revolute joint. For joints that are absent due to structural differences across designs, zero-range limits are used as placeholders to maintain a consistent parameter dimensionality.

Overall, this procedure maps diverse URDF descriptions into the unified canonical parameter set by combining mesh-based geometry estimation, joint-origin analysis, coordinate frame reorientation, and direct extraction of URDF-specified joint axes and limits. In practice, the resulting parameters typically require only a brief manual inspection and, when necessary, minor adjustments to ensure consistency with the canonical conventions.

\subsubsection{Canonical URDF Generation}
Reconstruction of a full URDF from canonical parameters is implemented using a template-based generation module built with the Jinja2 dynamic templating language~\cite{jinja}. The URDF is defined as a parameterized text template whose placeholders are populated with canonical parameter values.

Conditional logic within the template allows dynamic inclusion or omission of elements. For example, a joint is instantiated only if its lower and upper limits differ, and optional fingers or links are generated only when corresponding parameters are present. This design enables automatic generation of valid and consistent URDFs for hands with varying numbers of fingers, links, and joint configurations, while strictly adhering to the canonical conventions.

Together, the parsing and generation components provide a consistent, bidirectional conversion pipeline between diverse robotic hand models and the unified canonical representation, supporting both analysis of existing designs and synthesis of new hand embodiments.

\subsection{Morphology Latent Learning}  \label{app:morp}

\subsubsection{Data Sampling}
We first sample a global joint configuration for each synthetic hand, i.e., which fingers and which joints are present. Given this discrete topology, continuous geometric and frame parameters are sampled from uniform distributions over physically plausible ranges to preserve realistic hand proportions. Joint axes are drawn from six canonical directions $(\pm x, \pm y, \pm z)$ and encoded as one-hot vectors. Joint ranges are not supplied as continuous values; instead, each of the 22 canonical joints is represented by a binary indicator denoting its presence or absence.

Each sampled hand is serialized into a fixed-length vector by concatenating continuous geometry and frame parameters, joint-axis one-hot encodings, and joint-activation indicators. This unified representation provides the VAE with both continuous morphological variation and discrete structural information while keeping the sampling process simple and scalable.

\subsubsection{Model Architecture}
We use a standard variational autoencoder with MLP-based encoder and decoder networks. The encoder maps the input vector through three fully connected layers with hidden dimensions $[512, 256, 128]$, followed by BatchNorm and ReLU activations, and outputs the mean and log-variance of a 16-dimensional latent distribution. The decoder mirrors this architecture and reconstructs the hand parameters using multiple output heads: a continuous head for geometric and frame parameters, categorical heads for joint-axis prediction, and a sigmoid-activated head for joint-activation indicators.

\subsubsection{Training Objective}
Reconstruction losses are defined in a type-specific manner to reflect the heterogeneous nature of the parameters:
\vspace{-4pt}
\begin{equation}
    \small
    \begin{aligned}
        \mathcal{L}_{\text{cont}} &= \left\| \hat{q}_{\text{cont}} - q_{\text{cont}} \right\|_2^2, \\
        \mathcal{L}_{\text{axis}} &= \text{CrossEntropy}(\hat{q}_{\text{axis}}, q_{\text{axis}}), \\
        \mathcal{L}_{\text{joint}} &= \text{BinaryCrossEntropy}\!\left(\sigma(\hat{q}_{\text{joint}}), q_{\text{joint}}\right),
    \end{aligned}
\end{equation}
\vspace{-4pt}
where $q_{\text{cont}}$ denotes continuous morphology parameters, $q_{\text{axis}}$ the six-way joint-axis encodings, and $q_{\text{joint}}$ the binary joint-activation indicators. The overall loss combines these terms with a KL-divergence regularizer:
\begin{equation}
    \mathcal{L} =
    \mathcal{L}_{\text{cont}} +
    \mathcal{L}_{\text{axis}} +
    \mathcal{L}_{\text{joint}} +
    \beta\, \mathcal{L}_{\text{KL}},
\end{equation}
where $\beta=0.01$ in all experiments.

Training is performed using the Adam optimizer with a learning rate of $1\mathrm{e}{-4}$, $(\beta_1,\beta_2)=(0.95,0.999)$, and a weight decay of $1\mathrm{e}{-6}$.

\subsection{In-hand Reorientation}  \label{app:inhand}
\subsubsection{RL Network Architecture}
The reinforcement learning policy is implemented using a multi-layer perceptron (MLP) with hidden layer dimensions of [512, 256, 128]. To capture temporal dependencies across consecutive observations, a single gated recurrent unit (GRU) layer with a hidden size of 256 is applied before the MLP backbone. The GRU-processed observation is then concatenated with the original observation and passed as input to the MLP. The ELU activation function is used throughout the network.

\begin{figure}[t]
    \centering
    \begin{minipage}{0.5\columnwidth}
        \centering
        \captionsetup{width=0.95\linewidth}
        \captionof{table}{PPO training hyper-parameters.}
        \resizebox{\linewidth}{!}{
            \begin{tabular}{ll}
                \toprule
                \textbf{Hyper-parameter} & \textbf{Value} \\
                \midrule
                Discount factor $\gamma$ & 0.99 \\
                GAE $\lambda$ & 0.95 \\
                Horizon length $T$ & 32 \\
                Sequence length (RNN) & 4 \\
                Minibatch size & 32768 \\
                PPO epochs per update & 5 \\
                Learning rate & $5\times10^{-3}$ \\
                PPO Clip range $\epsilon$ & 0.2 \\
                KL threshold & 0.02 \\
                Entropy coefficient & 0.0 \\
                Critic loss coefficient & 4 \\
                Max gradient norm & 1.0 \\
                Reward scale & 0.01 \\
                Bounds loss coefficient & $1\times10^{-4}$ \\
                \bottomrule
            \end{tabular}
        }
        \label{tab:ppo-hparams}
    \end{minipage}
    \hfill
    \begin{minipage}{0.48\columnwidth}
        \centering
        \vspace{5pt}
        \includegraphics[width=0.9\linewidth]{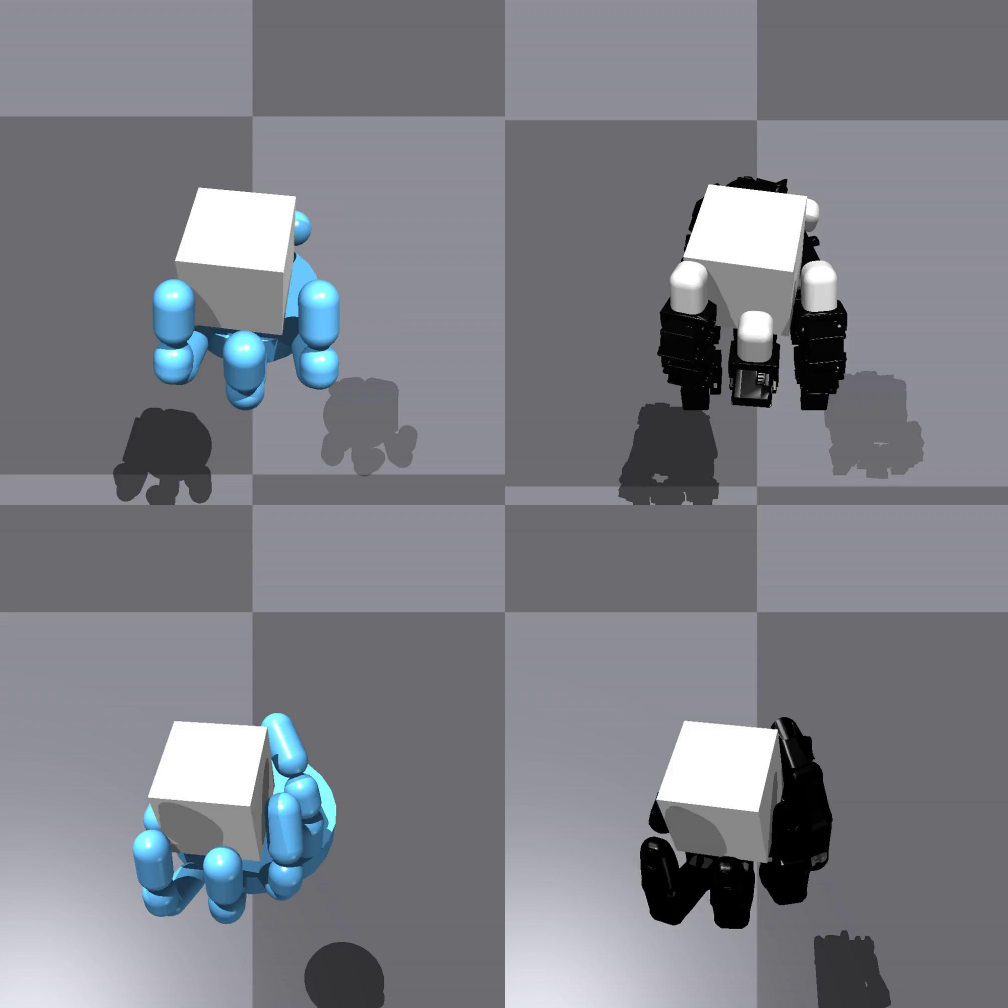}
        \captionsetup{width=0.92\linewidth}
        \caption{Visualization of in-hand reorientation under the original and canonical URDFs. Top: LEAP Hand; bottom: Shadow Hand.}
        \label{fig:inhand_vis}
    \end{minipage}
\end{figure}

\begin{table}[t]
    \centering
    \caption{Observation states used for policy training.}
        \begin{tabular}{cc}
            \toprule
            \textbf{Observation} & \textbf{Description} \\
            \midrule
            $\tilde{\mathbf{q}} \in \mathbb{R}^{n_\text{dof}}$ 
            & Normalized hand joint positions \\[4pt]
            
            $\mathbf{q}^{\text{tar}} \in \mathbb{R}^{n_\text{dof}}$ 
            & Normalized target joint (action) \\[4pt]
            
            $\mathbf{p}_\text{cube} \in \mathbb{R}^3$ 
            & Cube position \\[4pt]
            
            $\mathbf{r}_\text{cube} \in \mathbb{R}^3$ 
            & Cube orientation Euler angles. \\[4pt]
            
            $\dot{\mathbf{q}} \in \mathbb{R}^{n_\text{dof}}$ 
            & Hand joint velocities. \\[4pt]
            
            $\mathbf{v}_\text{cube} \in \mathbb{R}^3$ 
            & Cube linear velocity. \\[4pt]
            
            $\boldsymbol{\omega}_\text{cube} \in \mathbb{R}^3$ 
            & Cube angular velocity. \\[4pt]
            
            $\boldsymbol{\phi} \in \mathbb{R}^2$ 
            & Phase variables (periodic task encoding). \\
            \bottomrule
        \end{tabular}
    \vspace{-12pt}
    \label{tab:observations}
\end{table}

\begin{figure*}[t]
    \centering
    \begin{minipage}[t]{0.58\textwidth}
        \centering
        \includegraphics[width=\textwidth]{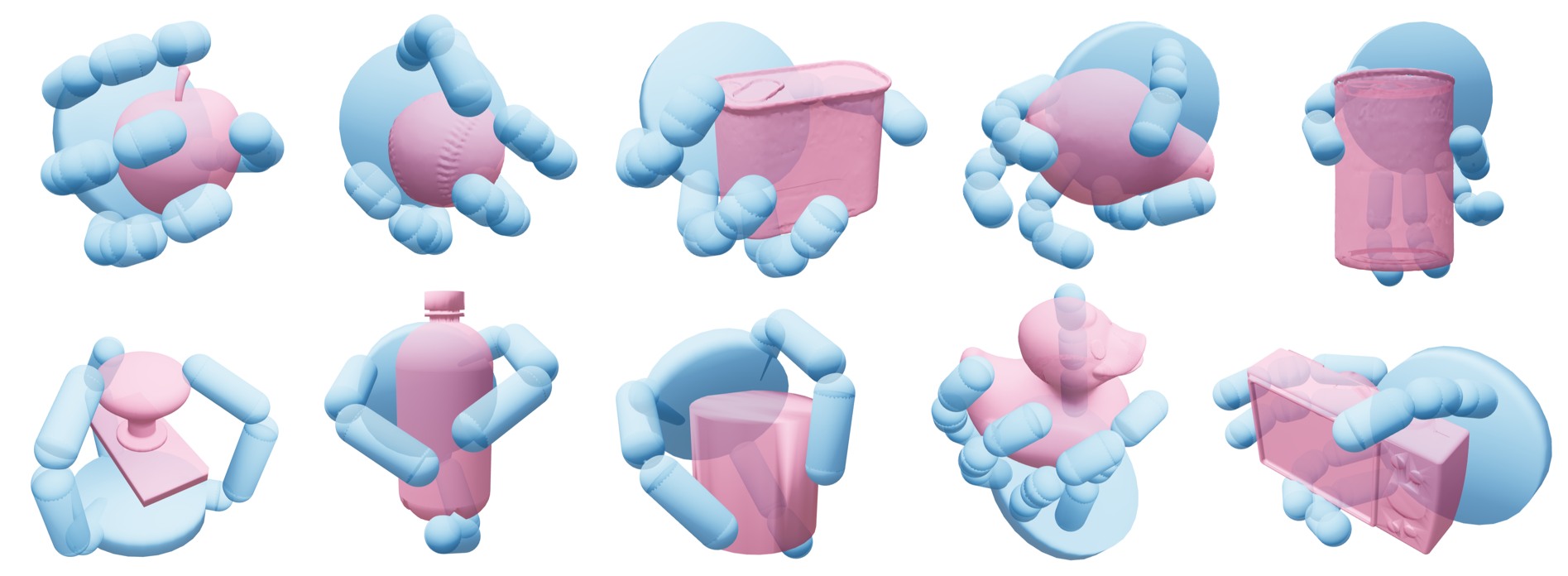}
        \caption{Grasp visualizations of the canonical URDF for the Allegro, Barrett, and Shadow Hand. All grasps are generated using the same cross-embodiment policy.}
    \end{minipage}
    \hfill
    \begin{minipage}[t]{0.38\textwidth}
        \centering
        \includegraphics[width=\textwidth]{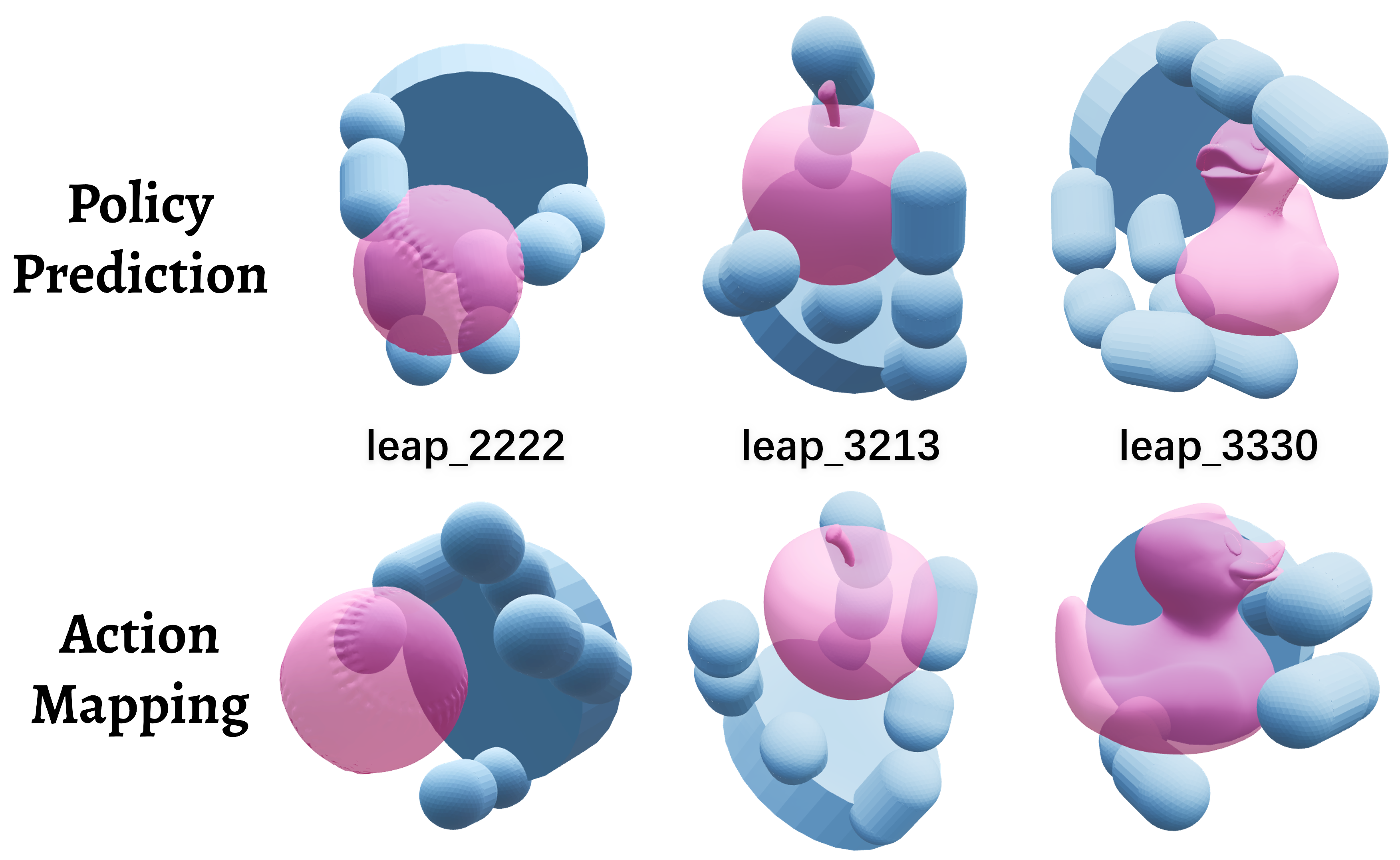}
        \caption{Grasp visualization comparison between direct inference on a target hand and mapping actions from \texttt{leap\_3333}.}
    \end{minipage}
    \vspace{-15pt}
\end{figure*}

\subsubsection{PPO Optimization Details}
We train our in-hand rotation policies using Proximal Policy Optimization (PPO) \cite{schulman2017proximal}. The reinforcement learning hyperparameters are summarized in Table \ref{tab:ppo-hparams}, and the observation inputs for the agents are listed in Table \ref{tab:observations}. $n_\text{dof}$ is the degree of freedom of the robot hand, and the parameters are the same for both the original robot hand and the canonical robot hands. We train the LEAP Hand policies for 400 gradient iterations (approximately 200M environment steps) and the Shadow Hand policy for 1,000 policy update iterations (approximately 500M environment steps). The reward function consists of the following reward and penalty components:

\begin{enumerate}
    \item $r_\text{rot} = \text{clip}(\omega_\text{z}, \omega_\text{min}, \omega_\text{max})$, where $\omega_\text{z}$ is the z-axis angular velocity of the cube, and the value is clipped between $\omega_\text{min}$ and $\omega_\text{max}$.
    \item $p_\text{pose} = \lVert q - q^0 \rVert_2^2$, where $q \in \mathbb{R}^{n_\text{dof}}$ is the current joint positions, and  $q^0 \in \mathbb{R}^{n_\text{dof}}$is the initial joint positions at the beginning of the episode.
    \item $p_\tau = \lVert \tau \rVert_2^2 $, where $\tau \in \mathbb{R}^{n_\text{dof}}$ is the applied torques of the robot joints.
    \item $p_\text{work} = (\tau^\top \dot{q})^2$ penalizes excessive force applied on the cube by the fingers.
    \item $p_\text{v} = \lVert \mathbf{v} \rVert_1$, where $\mathbf{v} \in \mathbb{R}^3$ is the linear velocity of the cube.
    \item The cube fell penalty is: 
    \begin{equation}
    r_{\text{fallen}} = 
    \begin{cases}
        1, & \text{if } z_\text{cube} < z_\text{threshold}, \\
        0, & \text{otherwise}.
    \end{cases}
\end{equation} where $z_\text{cube}$ is the z coordinate of the cube, and $z_\text{threshold}$ is height threshold to keep the cube from falling.
\end{enumerate}
 
The final reward is the aggregation of the individual reward terms, weighted by each term's scaling terms: 
$$
r_\text{base}
= s_\text{rot}r_\text{rot}
  - s_\text{v}\, p_\text{v}
  - s_\text{pose}\, p_\text{pose}
  - s_\tau\, p_\tau
  - s_\text{work}\, p_\text{work}.
$$

\subsubsection{Generate initial grasp configuration}
We construct our in-hand rotation task following previous works on in-hand rotation \cite{shaw2023leap, qi2023hand}. At the start of each episode, the hand is initialized in a stable grasp pose of a cube. To increase the diversity of initial configurations, we first generate a canonical stable grasp for each robot hand (e.g., LEAP Hand, Shadow Hand). We then introduce random perturbations to the joint angles to create varied but feasible grasp poses. For the LEAP Hand, joint angles are perturbed by uniformly sampled noise from $(-0.25,0.25)$ radians, while for the ShadowHand, we apply uniform noise from $(-0.1,0.1)$ radians. The perturbed configuration is then simulated for 50 steps without control inputs. During the rollout, small random forces are applied to the cube to perturb it further. At the end of each rollout, we validate the grasp using the following criteria: the fingertip–cube distance is below a threshold, at least two fingers are in contact with the cube, and the cube height exceeds a threshold relative to the palm center.

We generate 10,000 valid grasp configurations per robot hand, which serve as initial states. During training, environments randomly sample from these precomputed grasp poses at reset. For evaluation, we also randomly sample from the same set to ensure consistency across experiments.

\subsection{Cross-Embodiment Dexterous Grasping}  \label{app:grasp}

\subsubsection{Hyperparameters}
We use an MLP diffusion model with two hidden layers of sizes 512 and 256, and a diffusion-step embedding dimension of 64. The diffusion process is trained with 1000 timesteps using the ``sample'' prediction formulation. Optimization is performed with Adam using an initial learning rate of $1\mathrm{e}{-3}$ and a cosine annealing schedule that decays the learning rate to $1\mathrm{e}{-7}$ over the full training horizon.

\subsubsection{Evaluation Metric}  \label{app:grasp_metric}
We follow the evaluation procedure of $\mathcal{D(R,O)}$ Grasp~\cite{wei2024dro}, adapted to our canonical URDF controller. Grasp success is assessed using a force-closure–based criterion in Isaac Gym. For each predicted grasp, we execute the controller on the canonical hand and then sequentially apply external forces along the six orthogonal directions for 1 second each. A grasp is deemed successful if the resulting object displacement remains below 2 cm after all perturbations.

\begin{figure*}[t]
    \centering
    \resizebox{\linewidth}{!}{
        \includegraphics[width=\textwidth]{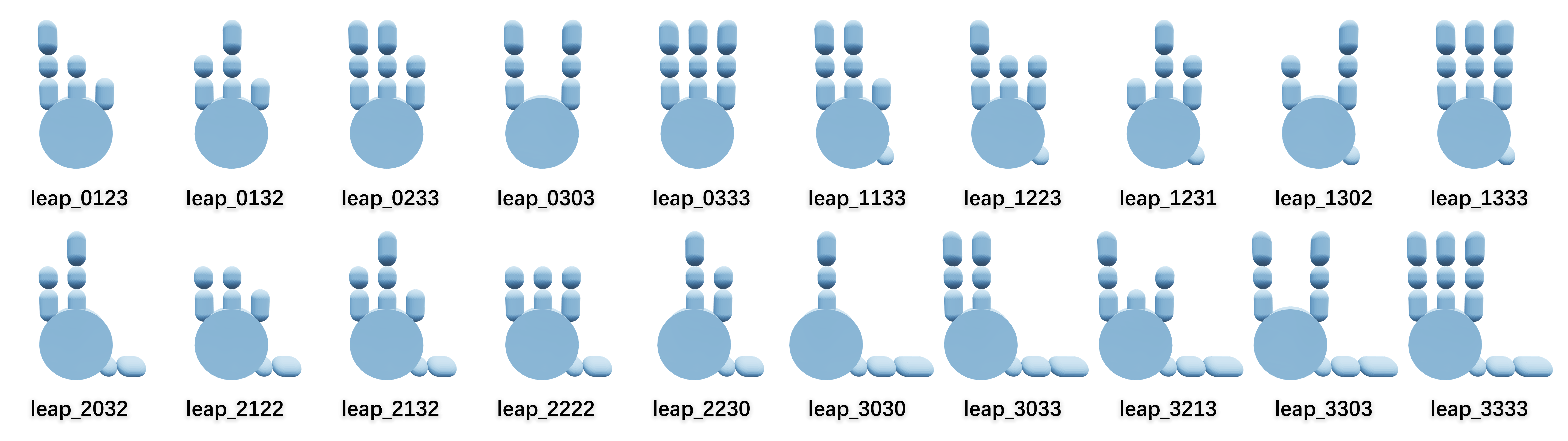}
    }
    \vspace{-16pt}
    \caption{Visualization of canonical LEAP Hand variants.}
    \label{tab:leap_variant}
\end{figure*}

\setlength{\tabcolsep}{2pt}
\begin{table*}[t]
    \centering
    \scriptsize
    \resizebox{\linewidth}{!}{
        \begin{tabular}{cccccccccccccccc}
        \toprule
        Hand & \# Grasps & Hand & \# Grasps & Hand & \# Grasps & Hand & \# Grasps &
        Hand & \# Grasps & Hand & \# Grasps & Hand & \# Grasps & Hand & \# Grasps \\
                \midrule
        \texttt{leap\_0000} &     0 & \texttt{leap\_0001} &     0 & \texttt{leap\_0002} &     0 & \texttt{leap\_0003} &     0 & \texttt{leap\_0010} &     0 & \texttt{leap\_0011} &     0 & \texttt{leap\_0012} &     0 & \texttt{leap\_0013} &     0 \\
        \texttt{leap\_0020} &     0 & \texttt{leap\_0021} &   157 & \texttt{leap\_0022} &   134 & \texttt{leap\_0023} &   304 & \texttt{leap\_0030} &     0 & \texttt{leap\_0031} &   452 & \texttt{leap\_0032} &   226 & \texttt{leap\_0033} & 25599 \\
        \texttt{leap\_0100} &     0 & \texttt{leap\_0101} &     0 & \texttt{leap\_0102} &     0 & \texttt{leap\_0103} &    65 & \texttt{leap\_0110} &     0 & \texttt{leap\_0111} &     0 & \texttt{leap\_0112} &     1 & \texttt{leap\_0113} &     4 \\
        \texttt{leap\_0120} &     0 & \texttt{leap\_0121} &     0 & \texttt{leap\_0122} &    34 & \texttt{leap\_0123} &   173 & \texttt{leap\_0130} &   640 & \texttt{leap\_0131} &   118 & \texttt{leap\_0132} &    13 & \texttt{leap\_0133} &  3801 \\
        \texttt{leap\_0200} &     0 & \texttt{leap\_0201} &   562 & \texttt{leap\_0202} &  2103 & \texttt{leap\_0203} & 10114 & \texttt{leap\_0210} &     9 & \texttt{leap\_0211} &    63 & \texttt{leap\_0212} &   212 & \texttt{leap\_0213} &   592 \\
        \texttt{leap\_0220} &   199 & \texttt{leap\_0221} &    61 & \texttt{leap\_0222} &  1567 & \texttt{leap\_0223} &  3293 & \texttt{leap\_0230} &  5972 & \texttt{leap\_0231} &  1010 & \texttt{leap\_0232} &   656 & \texttt{leap\_0233} & 11681 \\
        \texttt{leap\_0300} &     0 & \texttt{leap\_0301} &  3253 & \texttt{leap\_0302} & 20007 & \texttt{leap\_0303} & 49451 & \texttt{leap\_0310} &   514 & \texttt{leap\_0311} &  1401 & \texttt{leap\_0312} &  4171 & \texttt{leap\_0313} &  2615 \\
        \texttt{leap\_0320} &  6779 & \texttt{leap\_0321} &  1273 & \texttt{leap\_0322} & 12250 & \texttt{leap\_0323} & 22695 & \texttt{leap\_0330} & 54759 & \texttt{leap\_0331} &  4033 & \texttt{leap\_0332} &  6550 & \texttt{leap\_0333} & 28501 \\
        \texttt{leap\_1000} &     0 & \texttt{leap\_1001} &     0 & \texttt{leap\_1002} &     0 & \texttt{leap\_1003} &     7 & \texttt{leap\_1010} &     0 & \texttt{leap\_1011} &     0 & \texttt{leap\_1012} &     0 & \texttt{leap\_1013} &     0 \\
        \texttt{leap\_1020} &     0 & \texttt{leap\_1021} &     0 & \texttt{leap\_1022} &     0 & \texttt{leap\_1023} &     3 & \texttt{leap\_1030} &     1 & \texttt{leap\_1031} &   722 & \texttt{leap\_1032} &  2992 & \texttt{leap\_1033} &   482 \\
        \texttt{leap\_1100} &     0 & \texttt{leap\_1101} &     0 & \texttt{leap\_1102} &     0 & \texttt{leap\_1103} &    90 & \texttt{leap\_1110} &     0 & \texttt{leap\_1111} &     0 & \texttt{leap\_1112} &     3 & \texttt{leap\_1113} &     9 \\
        \texttt{leap\_1120} &     0 & \texttt{leap\_1121} &     0 & \texttt{leap\_1122} &     2 & \texttt{leap\_1123} &    27 & \texttt{leap\_1130} &     2 & \texttt{leap\_1131} &   115 & \texttt{leap\_1132} &   172 & \texttt{leap\_1133} &   539 \\
        \texttt{leap\_1200} &     0 & \texttt{leap\_1201} &   380 & \texttt{leap\_1202} &   261 & \texttt{leap\_1203} &   469 & \texttt{leap\_1210} &     0 & \texttt{leap\_1211} &     0 & \texttt{leap\_1212} &   237 & \texttt{leap\_1213} &    56 \\
        \texttt{leap\_1220} &     0 & \texttt{leap\_1221} &   116 & \texttt{leap\_1222} &   127 & \texttt{leap\_1223} &   294 & \texttt{leap\_1230} &   124 & \texttt{leap\_1231} &   752 & \texttt{leap\_1232} &  1555 & \texttt{leap\_1233} &   925 \\
        \texttt{leap\_1300} &     6 & \texttt{leap\_1301} &   274 & \texttt{leap\_1302} &   551 & \texttt{leap\_1303} &   616 & \texttt{leap\_1310} &     4 & \texttt{leap\_1311} &    26 & \texttt{leap\_1312} &   174 & \texttt{leap\_1313} &    86 \\
        \texttt{leap\_1320} &    60 & \texttt{leap\_1321} &   119 & \texttt{leap\_1322} &   918 & \texttt{leap\_1323} &   771 & \texttt{leap\_1330} &   231 & \texttt{leap\_1331} &   504 & \texttt{leap\_1332} &  2480 & \texttt{leap\_1333} &  2210 \\
        \texttt{leap\_2000} &     0 & \texttt{leap\_2001} &    39 & \texttt{leap\_2002} &    35 & \texttt{leap\_2003} &   954 & \texttt{leap\_2010} &     6 & \texttt{leap\_2011} &    63 & \texttt{leap\_2012} &    26 & \texttt{leap\_2013} &   274 \\
        \texttt{leap\_2020} &  1334 & \texttt{leap\_2021} &    48 & \texttt{leap\_2022} &   263 & \texttt{leap\_2023} &   507 & \texttt{leap\_2030} &  4452 & \texttt{leap\_2031} &    36 & \texttt{leap\_2032} &  1180 & \texttt{leap\_2033} &  2997 \\
        \texttt{leap\_2100} &     0 & \texttt{leap\_2101} &    11 & \texttt{leap\_2102} &     8 & \texttt{leap\_2103} &    54 & \texttt{leap\_2110} &     1 & \texttt{leap\_2111} &     0 & \texttt{leap\_2112} &     0 & \texttt{leap\_2113} &    21 \\
        \texttt{leap\_2120} &   138 & \texttt{leap\_2121} &    26 & \texttt{leap\_2122} &    28 & \texttt{leap\_2123} &    38 & \texttt{leap\_2130} &    62 & \texttt{leap\_2131} &    41 & \texttt{leap\_2132} &   144 & \texttt{leap\_2133} &    84 \\
        \texttt{leap\_2200} &   921 & \texttt{leap\_2201} &   147 & \texttt{leap\_2202} &   669 & \texttt{leap\_2203} &  2823 & \texttt{leap\_2210} &   223 & \texttt{leap\_2211} &     9 & \texttt{leap\_2212} &   234 & \texttt{leap\_2213} &   370 \\
        \texttt{leap\_2220} &   688 & \texttt{leap\_2221} &    55 & \texttt{leap\_2222} &   199 & \texttt{leap\_2223} &   761 & \texttt{leap\_2230} &  1629 & \texttt{leap\_2231} &    37 & \texttt{leap\_2232} &   492 & \texttt{leap\_2233} &  1168 \\
        \texttt{leap\_2300} &  2861 & \texttt{leap\_2301} &   436 & \texttt{leap\_2302} &  1759 & \texttt{leap\_2303} &  4677 & \texttt{leap\_2310} &   654 & \texttt{leap\_2311} &    93 & \texttt{leap\_2312} &   602 & \texttt{leap\_2313} &   833 \\
        \texttt{leap\_2320} &  2700 & \texttt{leap\_2321} &   110 & \texttt{leap\_2322} &   576 & \texttt{leap\_2323} &  1219 & \texttt{leap\_2330} &  3139 & \texttt{leap\_2331} &   161 & \texttt{leap\_2332} &   673 & \texttt{leap\_2333} &  1533 \\
        \texttt{leap\_3000} &     0 & \texttt{leap\_3001} &  5445 & \texttt{leap\_3002} &   369 & \texttt{leap\_3003} &  6129 & \texttt{leap\_3010} &  9222 & \texttt{leap\_3011} &  1014 & \texttt{leap\_3012} &   335 & \texttt{leap\_3013} &  1409 \\
        \texttt{leap\_3020} &  4229 & \texttt{leap\_3021} &  4860 & \texttt{leap\_3022} &   265 & \texttt{leap\_3023} &  6974 & \texttt{leap\_3030} & 38495 & \texttt{leap\_3031} &  9973 & \texttt{leap\_3032} &   500 & \texttt{leap\_3033} & 30216 \\
        \texttt{leap\_3100} &  3546 & \texttt{leap\_3101} &  1035 & \texttt{leap\_3102} &   157 & \texttt{leap\_3103} &  2069 & \texttt{leap\_3110} &  1091 & \texttt{leap\_3111} &   899 & \texttt{leap\_3112} &    55 & \texttt{leap\_3113} &   717 \\
        \texttt{leap\_3120} &   909 & \texttt{leap\_3121} &  1328 & \texttt{leap\_3122} &    85 & \texttt{leap\_3123} &  1785 & \texttt{leap\_3130} &  4837 & \texttt{leap\_3131} &   801 & \texttt{leap\_3132} &    31 & \texttt{leap\_3133} &  2942 \\
        \texttt{leap\_3200} &  3015 & \texttt{leap\_3201} &  5421 & \texttt{leap\_3202} &  2069 & \texttt{leap\_3203} &  3426 & \texttt{leap\_3210} &  9830 & \texttt{leap\_3211} &   551 & \texttt{leap\_3212} &   527 & \texttt{leap\_3213} &   451 \\
        \texttt{leap\_3220} &  1641 & \texttt{leap\_3221} &  2657 & \texttt{leap\_3222} &   804 & \texttt{leap\_3223} &  1577 & \texttt{leap\_3230} &  2986 & \texttt{leap\_3231} &  3861 & \texttt{leap\_3232} &   998 & \texttt{leap\_3233} &  5550 \\
        \texttt{leap\_3300} & 25530 & \texttt{leap\_3301} & 12502 & \texttt{leap\_3302} &  7703 & \texttt{leap\_3303} & 30369 & \texttt{leap\_3310} & 20580 & \texttt{leap\_3311} &  1417 & \texttt{leap\_3312} &  1800 & \texttt{leap\_3313} &  3599 \\
        \texttt{leap\_3320} &  9924 & \texttt{leap\_3321} &  4284 & \texttt{leap\_3322} &  2364 & \texttt{leap\_3323} & 12005 & \texttt{leap\_3330} & 35586 & \texttt{leap\_3331} &  5397 & \texttt{leap\_3332} &  3098 & \texttt{leap\_3333} & 14006 \\
        \bottomrule
        \end{tabular}
    }
    \vspace{3pt}
    \caption{Number of valid grasps generated for different LEAP Hand variants after filtering.}
    \vspace{-12pt}
    \label{tab:leap_dataset}
\end{table*}

\subsection{LEAP Hand Zero-Shot Generalization}  \label{app:zeroshot}

\subsubsection{Extended Canonical URDF for LEAP Hand}
To reduce the geometric discrepancy between the canonical URDF and the original LEAP Hand URDF under different configurations, we adopt an extended canonical URDF design in this experiment. In particular, we use the extended parameter set described in Appendix~\ref{app:extend} and adjust the placement of the abduction/adduction joints for non-thumb fingers by relocating them to the second link joint. This modification more faithfully reflects the distinctive kinematic structure of the LEAP Hand while remaining fully compatible with our canonical parameterization framework. When conditioning the policy, we directly use the original LEAP Hand morphology parameters as the hand condition.

While it is possible to further expand the parameter set to more precisely capture all aspects of the LEAP Hand design, doing so would introduce hand-specific parameters that are not shared by most dexterous hands. Since the abduction/adduction joint placement is the primary structural deviation of the LEAP Hand from the canonical design, we do not incorporate this variation into the general canonical URDF. Instead, we apply this extension only in this experiment to minimize the sim-to-real gap and ensure a fair evaluation of zero-shot generalization.

Importantly, this adjustment does not alter the underlying learning framework and illustrates the extensibility of our paradigm: hand-specific kinematic features can be incorporated through targeted extensions without compromising the generality of the canonical representation.

\subsubsection{LEAP Hand Variant Generation}

Thanks to the semantic structure of the canonical parameters, different LEAP Hand morphologies can be generated programmatically by a simple scripting procedure that sets the corresponding link and joint parameters to zero. Using the extended canonical URDF template, we batch-generate LEAP Hand variants with different numbers of links for each finger. Visualizations of representative variants used in our experiments are shown in Fig.~\ref{tab:leap_variant}, using Viser~\cite{yi2025viser}.

\begin{figure*}[t]
    \centering
    \resizebox{\linewidth}{!}{
        \includegraphics[width=\textwidth]{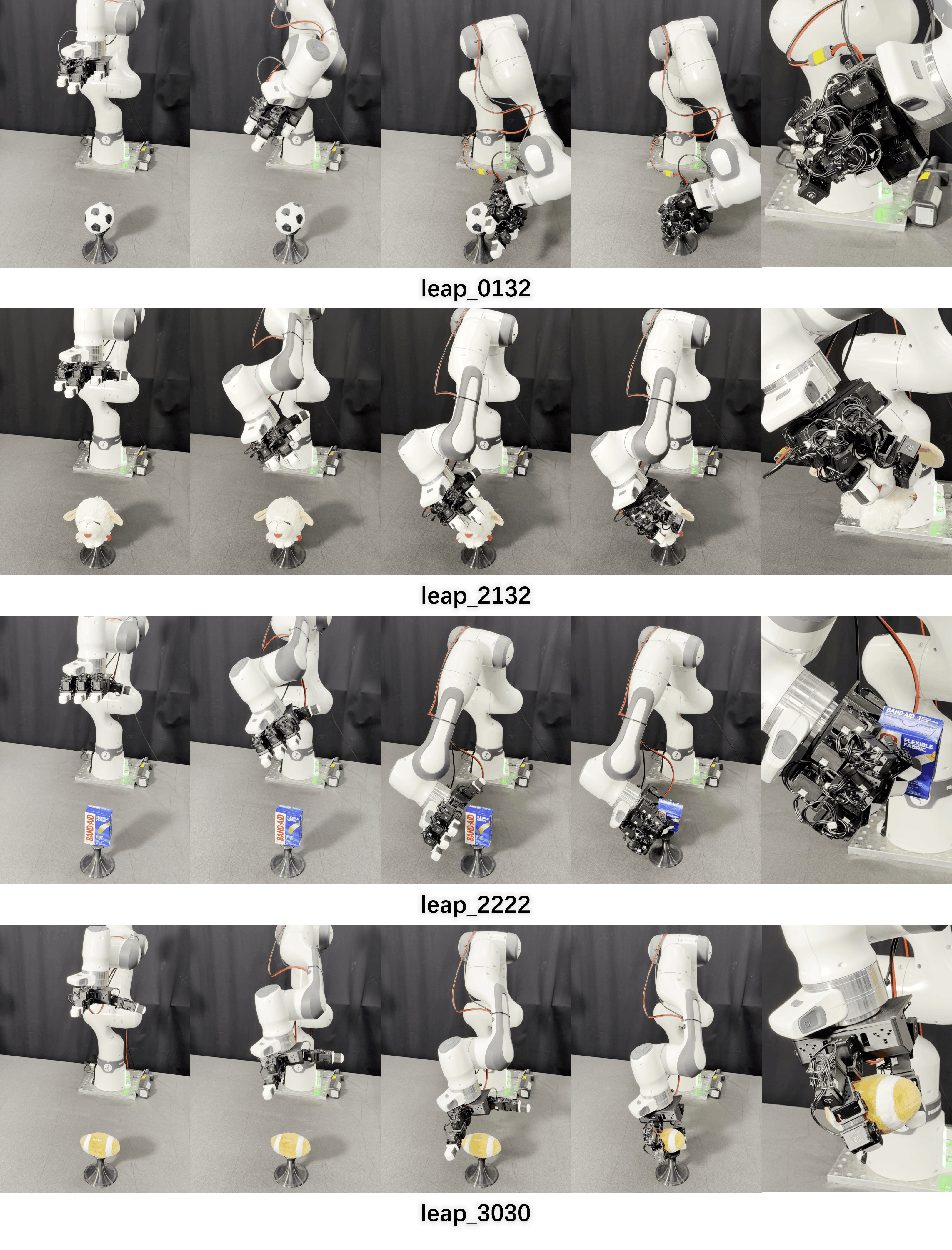}
    }
    \vspace{-20pt}
    \caption{Visualization of real-world experiment (I).}
    \label{tab:real_vis_1}
\end{figure*}

\begin{figure*}[t]
    \centering
    \resizebox{\linewidth}{!}{
        \includegraphics[width=\textwidth]{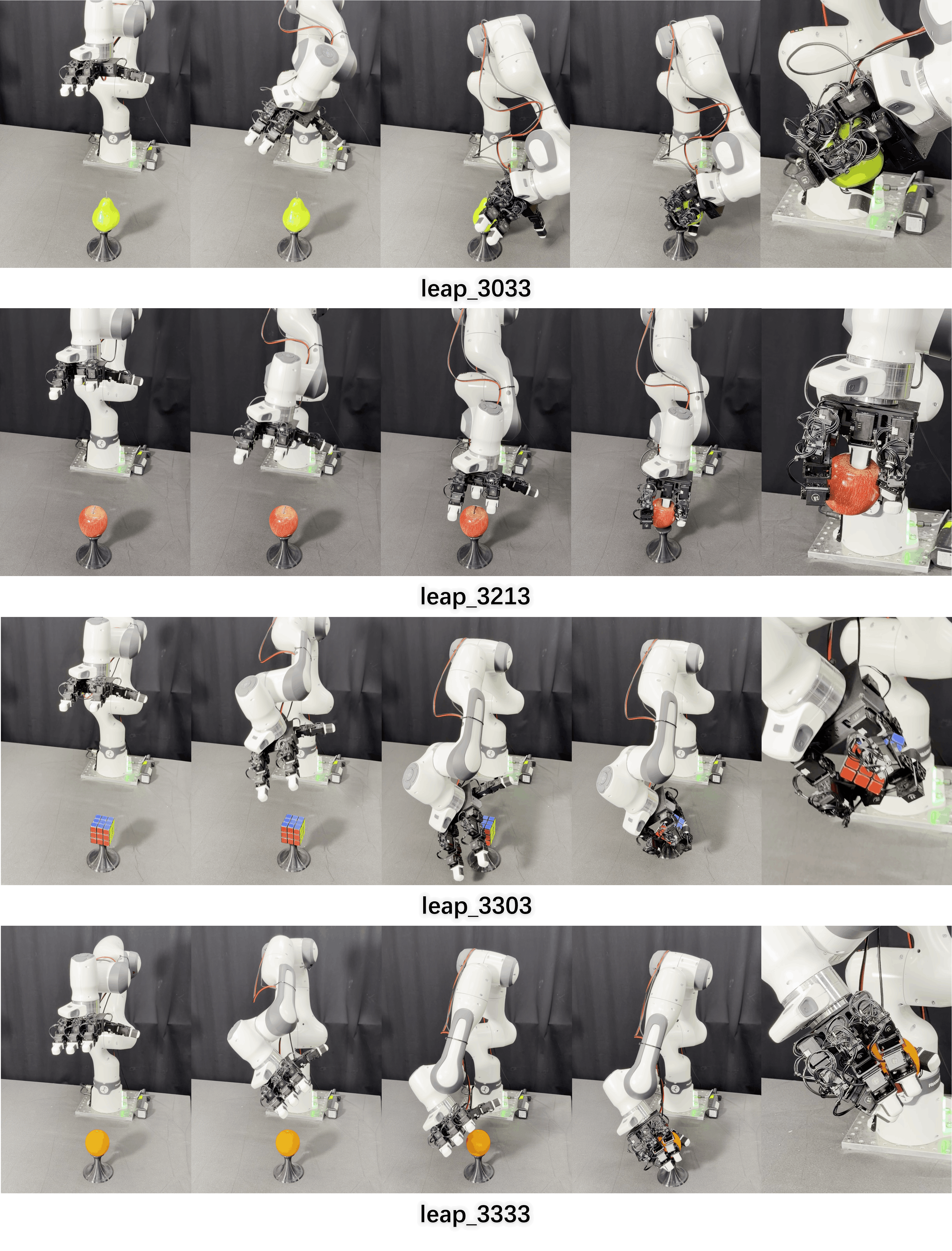}
    }
    \vspace{-20pt}
    \caption{Visualization of real-world experiment (II).}
    \label{tab:real_vis_2}
\end{figure*}

\subsubsection{Grasp Data Generation}

Grasp data for each LEAP Hand variant is generated using Lightning Grasp~\cite{yin2025lightning}, a recent analytical grasp synthesis method. For each hand morphology, we specify only the fingertip links and active joints in the configuration file, enabling efficient batch generation across all variants. We generate grasp candidates over four independent rounds to improve coverage.

All generated grasps are subsequently filtered in Isaac Gym using the same physical validity criteria as in the main experiments. Statistics of the resulting grasp datasets, including the number of valid grasps per hand variant, are summarized in Table~\ref{tab:leap_dataset}.

\end{document}